\documentclass[11pt]{article}

\usepackage{acl}
\usepackage{times}
\usepackage{latexsym}
\usepackage{comment}

\usepackage[T1]{fontenc}

\usepackage[utf8]{inputenc}

\usepackage{microtype}

\usepackage{inconsolata}

\usepackage{graphicx}
\usepackage{booktabs}

\title{BavGround: A Benchmark for Regional Cultural Grounding and \\Dialect Competence in Bavarian}

\author{
  \textbf{Jophin John\textsuperscript{1,*}}
  \quad
  \textbf{Michael Hoffmann\textsuperscript{1,*}}
\\
  \textbf{Jan Fillies\textsuperscript{2,3,*}}
  \quad
  \textbf{Michael A. Hedderich\textsuperscript{4,5}}
\\
  \textbf{Barbara Plank\textsuperscript{4,5}}
\\[0.7em]
  \textsuperscript{1}Leibniz Supercomputing Centre (LRZ), Garching, Germany
\\
  \textsuperscript{2}Stanford University, Stanford, California, USA
\\
  \textsuperscript{3}Freie Universität Berlin, Germany
\\
  \textsuperscript{4}Center for Information and Language Processing, LMU Munich
\\
  \textsuperscript{5}Munich Center for Machine Learning (MCML)
\\[0.4em]
  {\small \textsuperscript{*}Equal contribution.}
\\
 {\small \textbf{Correspondence:} \href{mailto:Jophin.John@lrz.de}{Jophin.John@lrz.de}, \href{mailto:Michael.Hoffmann@lrz.de}{Michael.Hoffmann@lrz.de}}
}

\newcommand{\projname}{\textsc{BavGround}}

\begin{document}
\maketitle

\begin{abstract}
Cultural evaluation of large language models (LLMs) often focuses on high-resource standard languages, leaving regional culture and dialect communities underrepresented. We introduce \projname, a benchmark for evaluating Bavarian regional cultural grounding and dialect competence across English, German and Bavarian. \projname{} contains 206 multiple-choice source questions across eight cultural domains per language, yielding 618 multi-parallel instances, with items covering both broadly accessible cultural knowledge and source-grounded regional knowledge from journalism, historical sources, and specialist literature. We evaluate fifteen 7B–10B open-weight instruction-tuned models and one closed-model reference. Strong multilingual models perform best overall, but performance drops on Bavarian items and source-grounded questions, indicating persistent difficulty with dialectal and localized cultural knowledge. We further show that conclusions depend strongly on evaluation protocol: raw answer-letter scoring, shuffled-letter scoring, option-text likelihood, generated-answer parsing, and semantic matching can produce different absolute scores and rankings, especially for regionally adapted models. Finally, an exploratory analysis of GENBA-10B checkpoints suggests that continued pretraining improves answer-content likelihood unevenly across domains, while dialect competence remains comparatively weak. \projname{} supports localized, protocol-aware evaluation of cultural representation in LLMs.

\end{abstract}

\section{Introduction}

Large language models (LLMs) are increasingly deployed in everyday settings, raising concerns about their reliability, fairness, and social impact \citep{bender2021dangers,blodgett2020language}. A growing body of work therefore evaluates whether LLMs represent culturally situated norms, values, linguistic practices, and forms of knowledge in reliable and equitable ways \citep{gallegos2024bias, liu2025culturally}. Recent cultural benchmarks have advanced multilingual and cross-cultural evaluation \citep{shi2024culturebank, myung2024blend, seveso2025italic, zhao2025makieval}, but many still operationalize culture at national levels, broad geographic regions, or high-resource standard languages. This leaves regional, dialectal, and minority-language communities comparatively underrepresented.

Regional cultures are not simply smaller versions of national identities: they may involve distinct histories, institutions, material practices, linguistic forms, and identity markers. Prior work emphasizes that cultural and linguistic evaluation should account for social context rather than treating language varieties as interchangeable surface forms \citep{blodgett2020language,hershcovich2022challenges}. Recent work suggests that dialect speakers can be disadvantaged by current LLMs, including in German dialect settings \citep{bui2025large}. These concerns motivate evaluation below the nation-state level, especially for regional varieties closely related to high-resource standard languages.

We study this problem through Bavarian, a regional language variety spoken across southern Germany, Austria, and northern Italy by approximately ten million people \citep{rowley2011bavarian}. Bavarian is a useful test case because it combines strong regional identity, substantial dialectal variation, and a close relationship to Standard German. We introduce \textbf{\projname}, a benchmark for Bavarian regional cultural grounding and dialect competence. \projname{} contains 206 source multiple-choice questions across eight thematic categories, translated into English, German, and Bavarian, yielding 618 evaluation instances. All German and Bavarian translations were produced manually by a native-speaking co-author. The benchmark combines broadly accessible general-knowledge questions with source-grounded regional questions derived from regional journalism, anthropological monographs, and specialist historical sources.

We evaluate fifteen instruction-tuned open-weight models in the 7B--10B range and find that \projname{} remains challenging even for strong multilingual systems: performance drops on Bavarian and on source-grounded regional questions. We further show that findings are highly sensitive to evaluation protocol. Canonical MCQA letter scoring can conflate cultural knowledge with answer-label priors, option order, and generation-format behavior, motivating our comparison with shuffled-label, option-text, generation-based, semantic-matching, and hidden-state diagnostics \citep{mizrahi2024state,wang2024my, dominguez2024questioning}. As an exploratory case study, we also study training dynamics and track \projname{} performance across continued-pretraining checkpoints of a Bavarian-centric model, GENBA-10B-it  \citep{hoffmann2025llama}, finding uneven gains across domains, languages, and protocols.

This paper contributes: (1) \projname, a manually validated benchmark for Bavarian regional cultural grounding and dialect competence across English, German, and Bavarian; (2) an evaluation of fifteen 7B--10B open-weight instruction-tuned models showing persistent difficulty on Bavarian and source-grounded regional questions; and (3) a protocol-aware analysis showing that single-protocol MCQ scores can obscure important model behaviors. We additionally provide an exploratory GENBA-10B checkpoint analysis.

\paragraph{Artifact Availability.}
To support transparency and reproducibility, we plan to release the benchmark,
evaluation code, generated outputs, and the interactive analysis dashboard (Appendix~\ref{app:dashboard}) upon
acceptance of the paper. For the review process, these materials are available
in anonymized form at:
\url{https://anonymous.4open.science/r/BavGround-7C68/README.md}.

\section{Related Work}
\subsection{Cultural Evaluation in NLP}
Research on cultural representation in NLP has expanded rapidly, particularly in the context of multilingual LLMs, fairness, and socially grounded generation \citep{gallegos2024bias, liu2025culturally}. Existing work studies culture through at least two complementary perspectives.

A first line of work conceptualizes culture through shared norms, values, beliefs, and socially situated behaviours, evaluating whether model outputs align with culturally specific expectations surrounding morality, religion, politeness, or social interaction \citep{ma-etal-2024-potential}. For example, \citet{naous2024having} study culturally situated continuations in religious and social contexts, while broader work examines how language models reproduce culturally dependent stereotypes \citep{abid2021persistent, kirk2021bias, kumar2025no}.

A second line of work conceptualizes culture through geographically situated knowledge, institutions, historical memory and tales \citep{bhagat2026tales, rooein2025biased, hedderich2025s}. This perspective is especially common in benchmark construction, where cultural competence is operationalized through factual or contextual question answering. Recent benchmarks evaluate knowledge associated with specific national or linguistic communities, including Italian cultural knowledge in ITALIC \citep{seveso2025italic} and multilingual cultural-awareness in MakiEval \citep{zhao2025makieval}, while broader cross-cultural NLP work highlights the challenges of representing diverse communities within current systems \citep{hershcovich2022challenges}.

Recent work has also questioned the theoretical assumptions underlying cultural NLP evaluation. \citet{zhou2025culture} argue that much current work relies on coarse cultural proxies, particularly nation-state boundaries, that fail to capture substantial within-community variation, proposing \emph{localization} as a more useful framing than static national categories. This critique is especially relevant for regional and minority language varieties, which remain comparatively underrepresented in existing benchmarks despite their distinct histories, institutions, and linguistic practices. \projname{} contributes to this literature by focusing on Bavarian as a regional variety with a substantial speaker population and strong regional identity. 

\subsection{Evaluation Protocols for Cultural and Multiple-Choice Benchmarks}
Existing work on cultural evaluation differs not only in what forms of culture is studied, but also in how cultural competence is measured. Common approaches include multiple-choice questionnaires in which models generate answer letters or score candidate options \citep{seveso2025italic, zhao2025makieval}, as well as generated continuations or free-form responses in culturally situated scenarios.

Recent work has increasingly shown that evaluation outcomes depend strongly on methodology. Prompt wording, answer ordering, parsing strategy, and decoding configuration can substantially affect measured performance \citep{mizrahi2024state}, and in multilingual settings, prompting language alone can alter measured cultural competence \citep{zhao2025makieval}. More broadly, \citet{wang2024my} demonstrate substantial mismatches between first-token probability rankings and generated responses across answer selection, refusal behaviour, and prompt perturbations—suggesting that standard MCQA scoring captures only part of a model's effective answer behaviour. Related work has similarly documented option-order sensitivity and answer-label bias in MCQ evaluation \citep{pezeshkpour2024large,zheng2024large}, as well as ordering, labeling, prompt-perturbation, and response-generation-method effects in survey-style LLM evaluations \citep{dominguez2024questioning,rupprecht2025prompt,ahnert2025survey}. Together, these findings suggest that benchmark scores can partly reflect elicitation artifacts rather than underlying knowledge.

Different evaluation methods also probe different capabilities in some setups: probability-based scoring measures preference over candidates, free-form generation evaluates instruction following and answer realization, semantic matching focuses on meaning independent of formatting, and representation-level analyses probe internal model geometry. As a result, protocols may not produce identical rankings even on the same benchmark.

Our work addresses this through a protocol-aware evaluation framework that compares probability-based scoring, answer-order perturbation, generation-based evaluation, semantic similarity matching, and hidden-state alignment diagnostics. We position this not as evidence that cultural benchmarks are unreliable, but as a demonstration that conclusions about cultural competence can depend substantially on evaluation methodology.

\subsection{Continued Pretraining and Cultural Representation}
Continued pretraining (CPT) has become a widely used strategy for adapting pretrained language models to new domains and languages without retraining from scratch \citep{gururangan2020don}, and is increasingly relied upon to expand linguistic coverage while preserving previously learned capabilities \citep{choudhury2025llama, koto2025sherkala}.

Most prior work evaluates CPT through aggregate downstream task performance or domain specialization, with comparatively little attention to how culturally grounded knowledge evolves during training—particularly for regional or dialectal varieties. Cultural evaluations also typically analyze only final checkpoints, treating cultural competence as a static property of released models.

We extend this literature through an exploratory longitudinal analysis of GENBA-10B \citep{hoffmann2025llama}, a German-English-Bavarian trilingual model developed through continued pretraining. Rather than evaluating only the final checkpoint, we examine how different forms of culturally grounded knowledge evolve across intermediate checkpoints and whether cultural domains improve uniformly during Bavarian-focused adaptation.

\begin{table*}[t]
  \small
  \centering
   \begin{tabular}{@{}p{2.5cm}p{0.9cm}p{11.5cm}@{}}
    \toprule
    \textbf{Category} & \textbf{Type} & \textbf{Example question} \\
    \midrule
    Historical & GEN & Which dynasty ruled Bavaria for centuries and ended its rule with the fall of the monarchy in 1918? \\
      & GRD & Why was King Ludwig III forced to leave Munich in a rented car on the evening of 7 November 1918? \\[3pt]
   Politics  & GEN & Which party has governed Bavaria continuously since 1957? \\
      & GRD & What is Bavaria often accused of by media in northern Germany? \\[3pt]
    Living Traditions      & GEN & What is the Bavarian tradition of erecting a decorated pole called? \\
      & GRD & Why are figures like the \textit{Vogelfa\"{a}nger} and \textit{Ba\"{a}rentreiber} part of the Mittenwald Maschkera tradition? \\[3pt]
    Culinary       & GEN & What is Obatzda? \\
      & GRD & What does the Bavarian dish Wasserschnalzn consist of? \\[3pt]
    Building \& Sacred Heritage     & GEN & Which Bavarian castle commissioned by King Ludwig II is often nicknamed a ``fairy-tale castle''? \\
      & GRD & Why did visitor numbers at Schloss Neuschwanstein not increase dramatically in 2025 despite receiving UNESCO World Heritage status? \\[3pt]
    Landscape  & GEN & Which river flows through Munich and continues north toward the Danube? \\
      & GRD & Which Bavarian independent city grew the most in population between 2010 and 2020? \\[3pt]
    Arts \& Identity      & GEN & Which composer is strongly associated with Bayreuth because his operas are performed at the Bayreuth Festival? \\
      & GRD & How did Franz Herzog von Bayern contribute to the public access of art collections? \\
    Language      & GEN & What does the Bavarian expression \textit{,,Geh, h\"{o}r auf!''} most likely mean in context? \\
      & GRD & What is the meaning of the Bavarian expression  \textit{,,Hawadehre``}? \\
    \bottomrule
  \end{tabular}
  \caption{Example questions from \projname, illustrating one general (GEN) and one grounded
    (GRD) question per category. GEN questions target broadly available cultural knowledge;
    GRD questions are sourced from regional journalism and specialist monographs and probe
    knowledge unlikely to appear in standard LLM pretraining corpora.}
  \label{tab:biabav-examples}
\end{table*}

\section{\projname{} Benchmark}

\subsection{Dataset Construction}
\projname{} is a multiple-choice benchmark for evaluating cultural knowledge and cultural bias in Bavarian. The benchmark comprises 206 source questions distributed across eight thematic categories of Bavarian culture and identity with two types each. Within each category, the benchmark is divided into two structurally distinct types (subsets). The first type (GEN) contains broadly accessible general-knowledge questions targeting widely recognizable aspects of Bavarian culture. The second contains grounded questions (GRD) derived from regional journalism, anthropological monographs, and specialist historical sources, targeting culturally specific knowledge unlikely to appear frequently in standard LLM pretraining corpora.
The benchmark covers both ideational and material aspects of culture. Categories such as Living Traditions and Customs and Bavarian Dialect and Language probe shared norms, practices, and linguistic identity, while categories such as Building and Sacred Heritage, Landscape, and Culinary foreground material artifacts, physical environments, and everyday practices. Historical and Politics cover the institutional and political dimensions of Bavarian identity. See Table~\ref{tab:biabav-examples} for examples of questions from both subsets across all eight categories.




\subsubsection{General Knowledge Questions}
The first ten questions in each category target broadly accessible cultural knowledge likely to appear in general multilingual training data. These questions were generated using Claude Sonnet 4.6 with category-specific prompts of the form: ``Create 10 multiple-choice questions that evaluate knowledge about Bavarian [category].'' The resulting items cover foundational topics such as historical figures, landmarks, culinary traditions, and characteristic features of Bavarian dialect and landscape, providing a baseline probe of general Bavarian cultural awareness in LLMs. All questions were subsequently checked for accuracy by two co-authors of the study and then translated into German and Bavarian by an in-house expert and native speaker.

\subsubsection{Grounded Questions}
The remaining questions in each category are grounded in specific primary sources selected manually by two in-house experts on a case-by-case basis from a combination of regional journalism and specialist monographs. Journalistic sources were drawn primarily from the Süddeutsche Zeitung and other regional Bavarian outlets (e.g. Traunsteiner Tagblatt); scholarly sources included anthropological monographs \cite{liu2021making, merlan2004preserving} and regional historical reference works. Sources were chosen with the explicit aim of identifying cultural knowledge that is unlikely to be well represented in the pretraining corpora of general-purpose LLMs: locally reported events, fine-grained ethnographic details, the specific institutional history of Bavarian political and religious life, and dialect-specific linguistic phenomena. Each grounded question is linked to its source document, with source metadata and URLs included in the dataset release; these sources also serve as references for answer validation. This subset is intended to be substantially more challenging than the general knowledge questions and to function as a targeted probe of whether LLMs possess culturally deep, regionally specific knowledge \cite{geertz2025thick, herzfeld2020poetics} rather than surface-level familiarity with Bavaria as a tourist destination.

\subsubsection{Distractor Construction}
Three distractor options per question were constructed manually and creatively by two co-authors with expert knowledge of Bavarian culture, history, and language, and native or near-native competence in English, German, and Bavarian. For grounded questions, distractors were additionally informed by the source documents used to construct the questions. All distractors were designed to be plausible but incorrect, requiring genuine cultural familiarity to distinguish from the correct answer, and were reviewed as part of the validation procedure described in Section 3.2.

\subsection{Annotation and Validation}
Validation was applied to both question subsets. For GEN questions, which were initially generated using Claude Sonnet 4.6, two in-house expert annotators with deep knowledge of Bavarian culture, history, and language independently reviewed each item for factual correctness, plausibility, discriminative quality, and clarity of the question stem. For GRD questions, which were constructed manually from regional sources, validation additionally covered the appropriateness of the assigned source. Each annotator reviewed the full set of questions produced by the other. Disagreements were resolved through discussion between the two annotators; items on which agreement could not be reached were removed from the dataset. Validation covered the original English questions and extended to the German and Bavarian translations.




\section{Experimental Setup}

\subsection{Models}
We evaluate fifteen instruction-tuned open-weight models in the 7B--10B range. For readability, we group them by family orientation rather than treating the groups as strict architectural classes: German/Bavarian-oriented models (GENBA-10B-it, Leo, LLaMmlein), European multilingual models (EuroLLM, Occiglot, Teuken, Pharia, Salamandra), and broader multilingual instruction models (Qwen2.5, Llama-3.1, Mistral, Gemma, OLMo, Granite, Aya). Appendix Table~\ref{tab:appendix-model-families} lists the exact model names used in the experiments.  We also include \texttt{gpt-5.4-mini} as a closed-model reference.


GENBA-10B \citep{hoffmann2025llama} is included both as a final instruction-tuned benchmark model and as a checkpoint case study. The checkpoint analysis evaluates 85 continued-pretraining checkpoints before final instruction tuning, from checkpoint 0 to 41{,}707; therefore, its final checkpoint is not expected to reproduce the instruction-tuned GENBA-10B result in Table~\ref{tab:model-comparison}.

\subsection{Experimental Design}
Our experiments use \projname{} as the central evaluation instrument in three ways. First, we evaluate fifteen instruction-tuned open-weight models on the full benchmark to characterize current performance on Bavarian regional cultural knowledge across languages, thematic categories, and GEN/GRD question types, and compare to a closed model (localized prompt templates are reported in Appendix~\ref{app:dataset-prompts}). Second, we test whether measured cultural competence is stable across evaluation protocols (See \ref{section4}).
Third, we apply \projname{}  longitudinally to 85 GENBA-10B checkpoints to study how regional cultural knowledge changes during a single continued-pretraining run. This checkpoint analysis is exploratory: it diagnoses domain-, language-, and protocol-specific changes over training, rather than establishing general causal laws about continued pretraining.


\subsection{Evaluation Framework}
\label{section4}
We evaluate \projname{} with a protocol-aware framework because MCQ scores can be affected by label priors, option order, tokenization, option length, and generation/parsing behavior. We compare three main views of the same benchmark items: \texttt{letter} measures answer-label likelihood by scoring A--D with conditional log-probability; \texttt{option\_text\_avg} measures answer-content likelihood by scoring each full option text and normalizing by token length; and \texttt{semantic\_embed\_generated\_answer} measures generated-answer meaning by mapping deterministic model outputs to the closest option using a multilingual Sentence-Transformers model based on MPNet \citep{reimers2019sentence,song2020mpnet}. We additionally run diagnostic variants, including shuffled-label scoring, unnormalized option-text scoring, generated-letter and generated-option parsing, question--option semantic baselines, and hidden-state alignment probes. These diagnostics are used to identify label priors, answer-format failures, shallow question--option similarity, and representation-level effects rather than as primary leaderboard metrics. Full protocol and implementation details are provided in Appendix~\ref{app:evaluation-artifacts}, Table~\ref{tab:appendix-protocol-details}.

\subsection{Metrics}
Our primary metric is accuracy, computed separately by protocol and reported by language, category, and GEN/GRD subset. For checkpoint analysis, we report accuracy trajectories across GENBA-10B checkpoints. For semantic matching and hidden-state alignment, we additionally inspect the correct-answer margin: similarity to the correct option minus the maximum similarity to any incorrect option. We use 95\% nonparametric bootstrap confidence intervals, computed by resampling benchmark items with replacement, for the main accuracy gaps tested in the Results: GEN versus GRD, English/German versus Bavarian, GENBA \texttt{option\_text\_avg} versus \texttt{letter}, and pairwise differences among the top three open-weight models under \texttt{letter} scoring. Details and intervals are reported in Appendix Table~\ref{tab:appendix-bootstrap-ci}. We include an interactive dashboard for model-, checkpoint-, language-, domain-, and strategy-level error analysis in the supplementary material; screenshots and JSON output details are in Appendix~\ref{app:dashboard}.

\section{Results and Analysis}
\label{Results}

\subsection{Open-Weight Models Struggle with Grounded and Dialectal \projname{} Items}
Table~\ref{tab:model-comparison} reports the standard \texttt{letter} scoring results for all fifteen instruction-tuned open-weight models and the closed-model reference. The strongest open-weight group consists of EuroLLM-9B-Instruct, Qwen2.5-7B-Instruct, and Llama-3.1-8B-Instruct, all clustered around 69\% accuracy. Their pairwise bootstrap intervals include zero, so we treat them as a tied top tier rather than as meaningfully separated. Granite-3.0-8B-Instruct follows closely at 66.3\%, while Occiglot-7B-Instruct and Aya-Expanse-8B form the next tier above 60\%. The open-weight mean is 53.0\%, indicating that \projname{} remains difficult for 7B to 10B open-weight models under standard multiple-choice letter scoring. The full category-expanded version of this table is provided in Appendix Table~\ref{tab:appendix-full-letter-model-comparison}, and additional cross-protocol model rankings are reported in Appendix Table~\ref{tab:appendix-model-metrics}.

The overall ranking also masks systematic language variation. Averaged across models, German is highest at 57.6\%, followed by English at 55.7\%, while Bavarian drops to 45.9\%. Bootstrap intervals confirm the Bavarian gap under \texttt{letter} scoring: English exceeds Bavarian by 9.8 points [7.5, 12.2], and German exceeds Bavarian by 11.7 points [9.5, 14.0]. This gap is visible even among the strongest models: EuroLLM reaches 74.3\% in German and 71.8\% in English, but 62.1\% in Bavarian; Llama-3.1 similarly falls from 75.2\% in German to 62.1\% in Bavarian. Qwen2.5 is the most balanced of the top models, with 72.8\% in English, 67.5\% in German, and 67.0\% in Bavarian. At the lower end, GENBA-10B-it reaches 41.9\% overall and LLaMmlein-7B-Chat is a clear outlier at 11.2\%, showing that standard letter selection is especially brittle for some regionally oriented or smaller instruction-tuned systems. Appendix Tables~\ref{tab:appendix-language-avg} and~\ref{tab:appendix-category-avg} give the corresponding language- and category-level averages across protocols.


As a closed-model reference point, \texttt{gpt-5.4-mini} reaches 89.6\% under first-token answer-letter log-probability scoring, 20.2 points above the strongest open-weight model, EuroLLM-9B-Instruct. Its performance is also more stable across languages, with 89.3\% in English, 91.3\% in German, and 88.3\% in Bavarian. The GEN--GRD gap remains visible even for this stronger model: accuracy drops from 95.0\% on GEN questions to 86.2\% on GRD questions. Thus, closed-model performance shows that \projname{} is solvable at high accuracy, but grounded and Bavarian-specific items still remain harder than broadly accessible cultural facts. Additional closed-model category breakdowns are shown in Appendix Table~\ref{tab:appendix-closed-model}.

\begin{table}[t]
\centering
\small
\setlength{\tabcolsep}{2.4pt}
\begin{tabular}{@{}lrrrrrr@{}}
\toprule
Model & All & EN & DE & BAV & GEN & GRD \\
\midrule
EuroLLM-9B     & 69.4 & 71.8 & 74.3 & 62.1 & 79.2 & 63.2 \\
Qwen2.5-7B     & 69.1 & 72.8 & 67.5 & 67.0 & 74.6 & 65.6 \\
Llama-3.1-8B   & 68.6 & 68.4 & 75.2 & 62.1 & 80.4 & 61.1 \\
Granite-3.0-8B & 66.3 & 72.3 & 62.1 & 64.6 & 72.5 & 62.4 \\
Occiglot-7B    & 61.8 & 55.8 & 70.9 & 58.7 & 77.5 & 51.9 \\
Aya-8B         & 60.8 & 64.6 & 71.4 & 46.6 & 73.3 & 52.9 \\
Mistral-7B     & 57.9 & 68.9 & 64.1 & 40.8 & 68.8 & 51.1 \\
OLMo-2-7B      & 54.2 & 64.1 & 48.5 & 50.0 & 60.4 & 50.3 \\
Salamandra-7B  & 52.1 & 51.9 & 63.1 & 41.3 & 66.7 & 42.9 \\
Gemma-2-9B     & 47.7 & 53.9 & 53.9 & 35.4 & 58.8 & 40.7 \\
Leo-7B         & 46.4 & 50.5 & 48.5 & 40.3 & 47.5 & 45.8 \\
Teuken-7B      & 44.5 & 44.7 & 51.9 & 36.9 & 59.6 & 34.9 \\
Pharia-7B      & 43.5 & 49.5 & 48.5 & 32.5 & 62.5 & 31.5 \\
GENBA-10B-it      & 41.9 & 35.9 & 50.0 & 39.8 & 55.8 & 33.1 \\
LLaMmlein-7B   & 11.2 & 10.2 & 13.6 & 9.7  & 8.8  & 12.7 \\
\midrule
Open-weight mean      & 53.0 & 55.7 & 57.6 & 45.9 & 63.1 & 46.7 \\
\midrule
\texttt{gpt-5.4-mini} & 89.6 & 89.3 & 91.3 & 88.3 & 95.0 & 86.2 \\
\bottomrule
\end{tabular}
\caption{Letter-scoring accuracy (\%) on \projname{} for open-weight models and the closed-model reference. \textit{All} is computed over all 618 translated instances; EN, DE, and BAV denote the three language subsets, while GEN and GRD denote general-knowledge and grounded subsets. Category-level and cross-protocol results are reported in Appendix Tables~\ref{tab:appendix-category-avg} and~\ref{tab:appendix-model-metrics}.}
\label{tab:model-comparison}
\end{table}


\subsection{General Knowledge Does Not Transfer Cleanly to Grounded Regional Knowledge}
\label{sec:gen-grd-results}

The letter-only breakdown in Table~\ref{tab:model-comparison} shows a consistent gap between general-knowledge (GEN) and grounded (GRD) questions. Across all models, average accuracy drops from 63.1\% on GEN to 46.7\% on GRD, a 16.4-point decrease with a 95\% bootstrap interval of [8.2, 24.3]. The strongest GEN score is Llama-3.1-8B-Instruct at 80.4\%, followed by EuroLLM at 79.2\% and Occiglot at 77.5\%. On GRD questions, Qwen2.5-7B-Instruct is strongest at 65.6\%, narrowly ahead of EuroLLM at 63.2\% and Granite at 62.4\%. This indicates that high performance on broadly accessible Bavarian cultural facts does not always transfer to source-grounded, region-specific questions.



Averaged over the fifteen open-weight models, \projname{} difficulty varies substantially by domain (Table~\ref{tab:main-category-protocols}; full results in Appendix Table~\ref{tab:appendix-category-avg}). Under \texttt{letter} scoring, Living Traditions, Arts \& Identity, and Politics are easiest, whereas Language is consistently hardest. This is not merely due to Bavarian prompt comprehension: Language remains difficult in English and German, indicating that reasoning about dialectal forms and meanings is a distinct challenge. Content-based scoring improves several domains, especially Historical and Building \& Sacred Heritage, but Language remains weak across protocols, suggesting that dialect competence remains difficult even when answer-label effects are reduced.

\begin{table}[t]
\centering
\small
\setlength{\tabcolsep}{1.8pt}
\renewcommand{\arraystretch}{0.92}
\begin{tabular}{@{}lrrrrrr@{}}
\toprule
& \multicolumn{3}{c}{All lang.} & \multicolumn{3}{c}{Let. by lang.} \\
\cmidrule(lr){2-4}\cmidrule(lr){5-7}
Category & Let. & Opt. & Sem. & EN & DE & BAV \\
\midrule
Arts \& Identity  & 61.2 & 61.9 & 67.0 & 63.7 & 65.9 & 54.1 \\
Building \& Sacred & 43.3 & 55.9 & 52.6 & 44.6 & 48.7 & 36.6 \\
Culinary   & 54.4 & 62.2 & 60.5 & 59.2 & 59.5 & 44.5 \\
History  & 54.3 & 68.1 & 57.2 & 55.2 & 58.9 & 48.8 \\
Landscape  & 54.3 & 63.5 & 57.2 & 58.7 & 59.5 & 44.8 \\
Language  & 35.6 & 39.4 & 45.3 & 37.1 & 41.6 & 28.3 \\
Traditions  & 61.4 & 63.3 & 60.5 & 65.7 & 63.7 & 54.8 \\
Politics   & 60.6 & 60.2 & 56.9 & 62.4 & 63.7 & 55.7 \\
\midrule
\multicolumn{7}{@{}l@{}}{\scriptsize Let.=letter; Opt.=option-text avg.; Sem.=semantic generated.} \\
\bottomrule
\end{tabular}
\caption{Average open-weight accuracy (\%) by category.}
\label{tab:main-category-protocols}
\end{table}


\subsection{Protocol Sensitivity Across Models: GENBA as a Case Study}

The preceding results show that \projname{} is challenging under standard \texttt{letter} scoring, but the broader protocol comparison shows that this is only one view of model behaviour. Across all fifteen models, accuracy rises from 53.0\% under \texttt{letter} to 59.3\% under \texttt{option\_text\_avg}; the corresponding option-text analogue of Table~\ref{tab:model-comparison} is reported in Appendix Table~\ref{tab:appendix-model-metrics}, with the full strategy breakdown in Appendix Table~\ref{tab:appendix-allmodel-strategies}.
This shift is model-specific: the strongest \texttt{letter} models remain stable or decrease slightly under \texttt{option\_text\_avg} (EuroLLM 69.4$\rightarrow$68.9, Qwen 69.1$\rightarrow$63.9, Llama 68.6$\rightarrow$64.9), whereas lower-ranked or strongly label-skewed models gain substantially (LLaMmlein +31.7, Pharia +18.9, GENBA +17.0). This suggests that protocol choice does not merely rescale the leaderboard: it reveals which models are robust across answer formats and which are penalized by answer-label or formatting effects. Label-prior and shuffled-label diagnostics are reported in Appendix Tables~\ref{tab:appendix-letter-label-prior-correlations} and~\ref{tab:appendix-letter-shuffled-diagnostic}.

GENBA-10B provides a clear case study of this protocol sensitivity. Although it is Bavarian-oriented, it ranks near the bottom under the default \texttt{letter} protocol. Under content-based evaluation, however, its performance is substantially higher: Table~\ref{tab:appendix-genba-protocol-summary} shows that GENBA increases from 41.9\% with \texttt{letter} scoring to 58.9\% with \texttt{option\_text\_avg} and 61.5\% with semantic matching of generated answers. The paired \texttt{option\_text\_avg}--\texttt{letter} gap is 17.0 points [10.7, 23.5] (Appendix Table~\ref{tab:appendix-bootstrap-ci}). We therefore interpret GENBA not as solving \projname{}, but as illustrating how answer-letter evaluation can understate recoverable answer content for some models. This recovery is also uneven across domains: the largest gains occur in \texttt{language} and \texttt{historical}, with smaller but still substantial gains in \texttt{building\_and\_sacred\_heritage} and \texttt{culinary} (Appendix Table~\ref{tab:appendix-genba-category-strategies}).

Importantly, alternative protocols do not erase \projname{}'s core difficulty structure: Bavarian remains below English and German under most strategies, and GRD questions remain harder than GEN questions. 

\begin{figure}[h!] 
    \centering
    \includegraphics[width=0.4\textwidth]{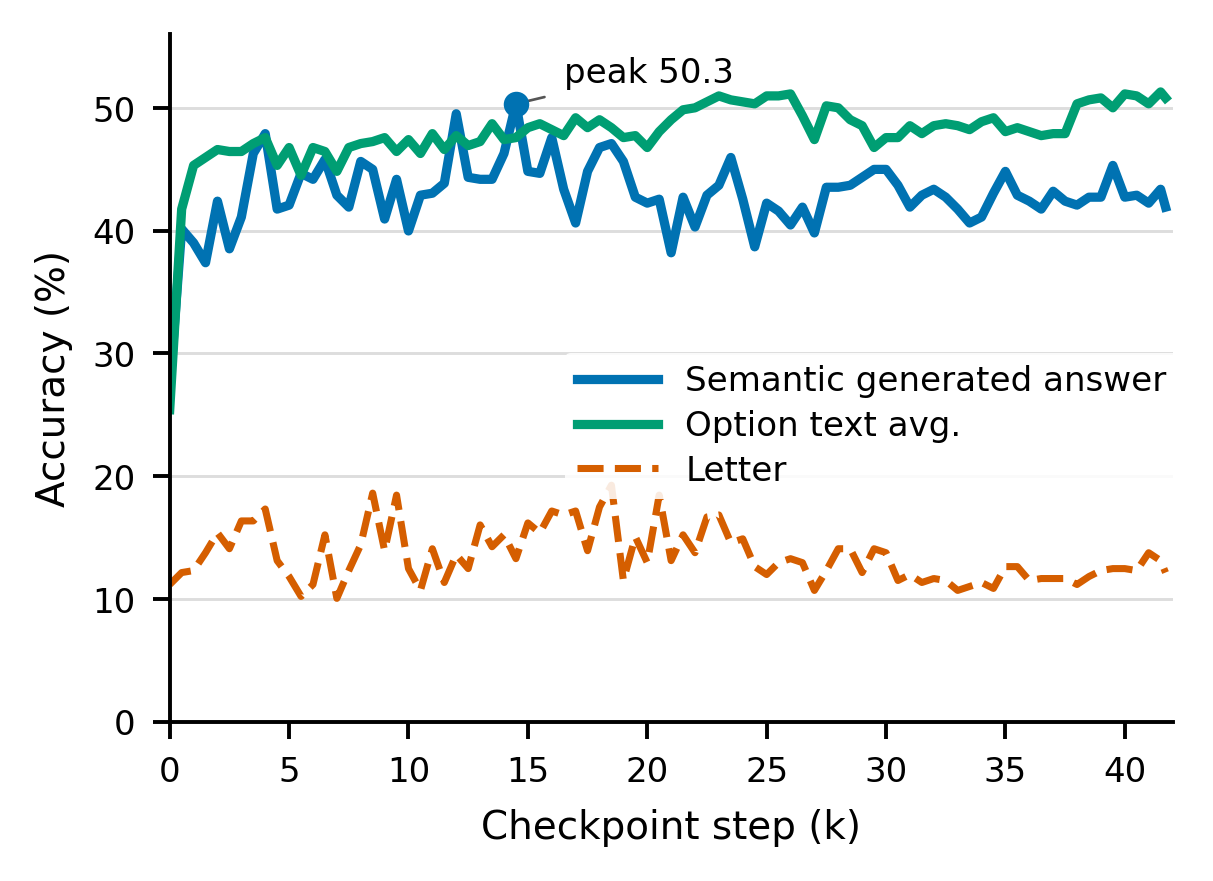}
    \caption{GENBA-10B continued-pretraining accuracy over all 618 instances before final instruction tuning.}
    \label{fig:checkpoint-trajectories}
\end{figure}

\subsection{Checkpoint Trajectories during Continued Pretraining}

We use the GENBA checkpoints to examine how regional cultural knowledge changes during continued pretraining. Figure~\ref{fig:checkpoint-trajectories} aggregates all English, German, and Bavarian instances and tracks checkpoints before final instruction tuning, so its final point is not expected to match the instruction-tuned GENBA row in Table~\ref{tab:model-comparison}. Across these checkpoints, \texttt{letter} accuracy remains low and unstable, rising from 11.2\% to a peak of 19.3\% before ending at 12.1\%. By contrast, \texttt{option\_text\_avg} improves from 25.4\% to 50.8\%, and \texttt{semantic\_embed\_generated\_answer} peaks at 50.3\% around checkpoint 14.5k. Continued pretraining therefore strengthens answer-content likelihood and generated-answer semantics more than answer-letter calibration.

The gains are uneven across languages and domains: Bavarian remains below English and German at the final checkpoint, and Language remains the weakest domain despite stronger improvements in Politics, Arts \& Identity, Living Traditions \& Customs, and Culinary. Other metrics reinforce this picture: semantic matching of generated answers improves substantially, while hidden-state alignment changes only weakly, indicating that better answer generation does not translate uniformly across all evaluation views. Full
breakdowns are provided in Appendix Tables~\ref{tab:appendix-checkpoint-metrics}--\ref{tab:appendix-checkpoint-language} and Appendix Figures~\ref{fig:checkpoint-strategy-trajectories}--\ref{fig:checkpoint-bar-category-strategy-trajectories}. Overall, the checkpoint analysis suggests uneven capability development rather
than uniform improvement across different domains.

\section{Discussion}
Our findings highlight three implications for evaluating culturally grounded knowledge in LLMs. First, regional cultural knowledge is not a single capability. Models perform better on broadly recognizable Bavarian facts than on dialect competence and source-grounded regional items. We do not claim that every grounded item is intrinsically ``deeper'' than encyclopaedic knowledge; some practices are well documented in public sources, as the all-model error examples in Appendix Table~\ref{tab:appendix-all-model-public-errors} illustrate. The point is methodological: national or tourism-facing coverage does not guarantee competence on localized formulations, dialect expressions, institutional details, or source-specific regional knowledge. \projname{} therefore tests whether models can move beyond recognizing Bavaria as a named cultural region toward answering questions across language variety, domain and evidence type.

Second, our results corroborate prior work showing that MCQA evaluations are protocol-sensitive, and extend this observation to regional cultural grounding. On \projname{}, protocol choice changes not only absolute scores but also the failure modes that become visible. Standard \texttt{letter} scoring can understate performance for models such as GENBA by conflating cultural knowledge with answer-label and position behaviour, while EuroLLM's drop under semantic matching shows that strong option likelihood does not necessarily imply reliable answer realization. We therefore use shuffled-letter, option-text, semantic, and hidden-state diagnostics to separate label priors, answer-content preference, generated-answer meaning, and representation-level alignment.

Finally, the checkpoint analysis is an exploratory case study of GENBA-10B. Continued pretraining improves some evaluation views substantially but unevenly: Bavarian remains below English and German, dialect remains the weakest domain, and hidden-state alignment does not consistently track option-text gains. BAVGROUND can therefore help diagnose whether systems need dialect-focused instruction tuning, answer-format calibration, or community-informed evaluation before deployment with Bavarian-speaking users.

\section{Conclusion and Future Work}

We presented \projname{}, a localized benchmark for evaluating Bavarian regional cultural grounding and dialect competence across English, German, and Bavarian. By combining general cultural questions with source-grounded regional items and multiple evaluation protocols, \projname{} enables controlled analysis of cultural knowledge below the level of national categories and high-resource standard languages.

Our experiments show that localized cultural evaluation should be both domain-aware and protocol-aware. The same benchmark surfaces several distinct phenomena: persistent difficulty on Bavarian and source-grounded items, variation across cultural domains, sensitivity to answer format, and uneven capability development during continued pretraining. These findings suggest that regional cultural competence is better understood as a set of interacting capabilities rather than as a single aggregate accuracy score.

\projname{} is a starting point for localized, protocol-aware cultural evaluation. Future work should extend it to other regional and minority language communities and broaden community participation.

\section{Limitations}
First, our evaluation covers fifteen instruction-tuned models in the 7B–10B parameter range, providing a broad cross-section of current open-weight model families but leaving larger-scale and closed-weight systems largely uncharacterized. The single closed-model reference (gpt-5.4-mini) contextualizes benchmark difficulty but is insufficient to draw conclusions about frontier model behavior. Findings about the relationship between model orientation and cultural performance may not generalize beyond this parameter range, and we encourage future work to evaluate \projname{} against larger open-weight and closed systems as they become accessible.

Second, the checkpoint analysis is restricted to GENBA-10B, meaning that the non-monotonic trajectories, domain-specific gains, and protocol divergences we document are properties of one continued-pretraining run under one training configuration. We cannot determine whether these dynamics, including the divergence between letter accuracy and option-text likelihood, or the persistent weakness of the Language domain, reflect general properties of Bavarian-focused adaptation or idiosyncrasies of GENBA's architecture, data mixture, or training schedule. Capacity-matched control runs trained on non-Bavarian data would be necessary to attribute observed checkpoint gains to cultural content exposure rather than general continued-pretraining effects, and we do not include such controls here.

Third, the ten general-knowledge questions per category were generated using Claude Sonnet 4.6. Since Claude model outputs are widely used in LLM post-training pipelines, some of the evaluated models may have been exposed to similar or overlapping content during instruction tuning, potentially inflating performance on the general-knowledge subset relative to the grounded questions. We cannot rule out this form of benchmark contamination, and future versions of \projname{} should consider replacing AI-generated general questions with human-authored items or questions drawn from independently verified sources.

Fourth, \projname{} was manually translated into English, German, and Bavarian, but Bavarian is not a single uniform variety: it encompasses considerable dialectal variation across much of Bavaria, Austria, and South Tyrol, and our translations target one written representation of this continuum. We do not calculate inter-translator agreement for the Bavarian translations specifically, and we do not evaluate whether the translated items preserve the pragmatic and dialectal specificity of the source questions. Given that dialect competence is a central claim of the benchmark, this is a meaningful gap. Future work should document translation methodology more rigorously, validate Bavarian items with native speakers from multiple dialectal sub-regions, and consider whether a single written Bavarian form is sufficient or whether sub-regional variation warrants separate evaluation tracks.

Fifth, the source-grounded questions are constrained by the availability of relevant anthropological literature: despite consulting domain experts, only one key monograph \citep{liu2021making} and one key article \citep{merlan2004preserving} on Bavarian culture could be identified as suitable primary sources. This scarcity also restricts the overall size of the grounded subset, as the manual creation of source-tied questions, each requiring careful reading, item construction, and expert review, is substantially more time-intensive than automated generation. Future work should seek to expand the range of primary sources consulted, potentially in collaboration with regional libraries, ethnographic archives, or Bavarian cultural institutions.

Sixth, the Bavarian translations were produced by a single translator who, although a native Chiemgau variety speaker, has resided outside Bavaria for an extended period. Prolonged absence from a dialect community is likely a driver of dialect attrition, and the translations may consequently exhibit a low Levenshtein distance from the corresponding German translations. Model performance on Bavarian items should therefore be interpreted as an upper bound: the true capability gap between models on German and Bavarian is likely larger than our results indicate.

Finally, our longitudinal design across 85 checkpoints does not support strong causal claims about the mechanisms driving observed changes in cultural representation. Disentangling increased model capacity from the influence of Bavarian-specific training data remains difficult \citep{betley2026training}, and training smaller models from scratch with intermediate checkpoints would provide a cleaner handle on these causal dynamics.

\section{Ethical Concerns}
In this work we used only publicly available models and source materials and did not collect or process any private or sensitive personal data. The evaluated open-weight models are used only as evaluation targets, and we do not redistribute third-party model weights. Their use is governed by the corresponding model cards, licenses, and API terms; Appendix Table~\ref{tab:appendix-model-families} reports the exact model identifiers or Hugging Face links where available. The closed-model reference is accessed only through its API terms.

Because cultural benchmarks necessarily simplify complex and internally diverse identities, \projname{} should not be interpreted as a definitive representation of Bavarian culture or dialect usage. We further caution against interpreting any single benchmark score as a definitive measure of cultural competence or bias, since our results show that evaluation protocol substantially affects measured performance. The generated model outputs and checkpoint behaviors analyzed in this work do not represent the personal views of the authors.

As part of our transparency and reproducibility commitments, we provide an
anonymized artifact for review and plan to release the benchmark, evaluation
code, and generated outputs upon acceptance.
\bibliography{custom}

\clearpage
\appendix

\section{Appendix Overview}
\label{app:overview}

This appendix follows the structure of the main paper. Appendix~\ref{app:dataset-prompts} documents the benchmark instances and prompt templates. Appendix~\ref{app:evaluation-artifacts} describes the result artifacts used for the analyses. Appendix~\ref{app:open-model-results} provides additional open-weight model results from \texttt{results\_allmodels}. Appendix~\ref{app:closed-model-results} reports the closed-model reference results from \texttt{results\_closedmodels}. Appendix~\ref{app:genba-protocol-diagnostics} expands the protocol-sensitivity analysis for GENBA-10B. Appendix~\ref{app:checkpoint-results} gives checkpoint-level summaries and figures from \texttt{results\_genba\_checkpoints}. Appendix~\ref{app:dashboard} describes the dashboard and stored JSON outputs.

\section{Dataset and Prompt Details}
\label{app:dataset-prompts}

The evaluated \projname{} files contain 206 source questions translated into English, German, and Bavarian, yielding 618 multiple-choice instances. Table~\ref{tab:appendix-dataset-counts} reports the category distribution used by the result files.

\begin{table}[h]
\centering
\small
\begin{tabular}{lrr}
\toprule
Category & Source Qs. & Instances \\
\midrule
Arts \& Identity & 25 & 75 \\
Building \& Sacred Heritage & 29 & 87 \\
Culinary & 25 & 75 \\
Historical & 25 & 75 \\
Landscape & 25 & 75 \\
Language & 25 & 75 \\
Traditions \& Customs & 27 & 81 \\
Politics & 25 & 75 \\
\midrule
Total & 206 & 618 \\
\bottomrule
\end{tabular}
\caption{\projname{} category counts. Each source question is evaluated in English, German, and Bavarian.}
\label{tab:appendix-dataset-counts}
\end{table}

Table~\ref{tab:appendix-all-model-public-errors} lists representative items missed by all evaluated systems under standard \texttt{letter} scoring, including the closed-model reference. The GRD examples qualify our interpretation of source-grounded difficulty: these items are linked to public sources in the dataset metadata, yet models still consistently select distractors. Their failure therefore cannot be explained simply by inaccessible evidence or lack of web presence. The GEN example shows that errors also occur for broader dialect knowledge when the question requires precise linguistic competence. We use these cases to support a limited methodological claim: public documentation or regional-media coverage does not by itself imply that models can answer localized cultural questions reliably.

\begin{table*}[t]
\centering
\scriptsize
\setlength{\tabcolsep}{3pt}
\resizebox{\textwidth}{!}{
\begin{tabular}{p{2.2cm}p{0.8cm}p{2.5cm}p{4.8cm}p{3.0cm}p{2.7cm}p{2.1cm}}
\toprule
Category & Subset & Item & Question focus & Gold answer & All-model distractor pattern & Public source \\
\midrule
Language & GEN & \texttt{language\_en\_003} & Bavarian sound shift for Standard German \textit{ei}, as in \textit{nein} & B: oa $\rightarrow$ noa & A: 15, D: 1 & General item; no external source \\
Building \& Sacred Heritage & GRD & \begin{tabular}[t]{@{}l@{}}\texttt{building\_and\_sacred}\\\texttt{\_heritage\_en\_013}\end{tabular} & Museum displaying a Maria Magdalena figure formerly from the Dominikanerkirche in Augsburg & B: Louvre & A: 15, D: 1 & \href{https://www.sueddeutsche.de/bayern/dominikanerkirche-augsburg-wiedereroeffnung-geschichte-li.3348154}{SZ article} \\
Historical & GRD & \texttt{historical\_en\_012} & First meeting place of the Worker, Soldier, and Farmer Council under Kurt Eisner on 7 Nov. 1918 & C: Landtag building in Prannerstraße & A: 15, D: 1 & \href{https://www.sueddeutsche.de/muenchen/100-jahre-freistaat-bayern-revolution-schauplaetze-1.4199230}{SZ article} \\
Language & GRD & \texttt{language\_en\_015} & Meaning of the Bavarian dialect word \textit{Stranitzl} & C: Eine Papiertüte & A: 8, B: 7, D: 1 & \href{https://www.sueddeutsche.de/quiz/bayern/himmelherrgottnoamoi-gpPDEP4}{SZ quiz} \\
Living Traditions \& Customs & GRD & \begin{tabular}[t]{@{}l@{}}\texttt{living\_traditions}\\\texttt{\_and\_customs\_en\_021}\end{tabular} & Year in which Mariä Lichtmess lost official public-holiday status & C: In 1912 & A: 9, D: 5, B: 2 & \href{https://www.sueddeutsche.de/bayern/bayern-mariae-lichtmess-bedeutung-feiertag-1.6342757}{SZ article} \\
\bottomrule
\end{tabular}
}
\caption{Examples of GEN and publicly sourced GRD items missed by all 15 open-weight models and the \texttt{gpt-5.4-mini} closed-model reference under standard \texttt{letter} scoring. Distractor counts report the predicted labels among the 16 systems; the gold label is omitted from the pattern because no system selected it. SZ denotes Süddeutsche Zeitung.}
\label{tab:appendix-all-model-public-errors}
\end{table*}

The localized answer-generation prompt asks the model to start with the selected option letter and optionally continue with the exact option text. The English template is:
\begin{quote}\small
\texttt{\{question\}}\\
\texttt{Options:}\\
\texttt{A: \{option\_A\}}\\
\texttt{B: \{option\_B\}}\\
\texttt{C: \{option\_C\}}\\
\texttt{D: \{option\_D\}}\\
\texttt{Reply by starting with the correct option letter: A, B, C, or D. You may optionally continue with the exact option text. Examples: 'B' or 'B: Lenbachhaus'.}\\
\texttt{Answer:}
\end{quote}

The German instruction is:
\begin{quote}\small
\texttt{Beginne mit dem richtigen Buchstaben A, B, C oder D. Danach darfst du optional den exakten Optionstext nennen. Beispiele: 'B' oder 'B: Lenbachhaus'.}
\end{quote}

The Bavarian instruction is:
\begin{quote}\small
\texttt{Fang mit dem richtigen Buchstaben A, B, C oda D o. Danach derfst optional den exakten Optionstext dazuschreibn. Beispui: 'B' oder 'B: Lenbachhaus'.}
\end{quote}

\section{Evaluation Artifacts and Protocols}
\label{app:evaluation-artifacts}

Table~\ref{tab:appendix-model-families} lists the open-weight model groupings used in the experimental setup together with the exact result-file target names and model links. These groupings are intended to make the model set readable; they are not mutually exclusive claims about pretraining data or architecture.

\begin{table*}[t]
\centering
\small
\setlength{\tabcolsep}{3pt}
\resizebox{\textwidth}{!}{
\begin{tabular}{p{4.6cm}p{5.4cm}p{6.4cm}}
\toprule
Grouping & Open weight model name & Model identifier / link \\
\midrule
German/Bavarian-oriented & \texttt{genba-10b-it} & local GENBA-10B instruction-tuned release; no Hugging Face link \\
German/Bavarian-oriented & \texttt{leo-hessianai-7b-chat} & \href{https://huggingface.co/LeoLM/leo-hessianai-7b-chat}{\texttt{LeoLM/leo-hessianai-7b-chat}} \\
German/Bavarian-oriented & \texttt{LLaMmlein\_7B\_chat} & \href{https://huggingface.co/LSX-UniWue/LLaMmlein_7B_chat}{\texttt{LSX-UniWue/LLaMmlein\_7B\_chat}} \\
European multilingual & \texttt{EuroLLM-9B-Instruct} & \href{https://huggingface.co/utter-project/EuroLLM-9B-Instruct}{\texttt{utter-project/EuroLLM-9B-Instruct}} \\
European multilingual & \texttt{occiglot-7b-eu5-instruct} & \href{https://huggingface.co/occiglot/occiglot-7b-eu5-instruct}{\texttt{occiglot/occiglot-7b-eu5-instruct}} \\
European multilingual & \texttt{Teuken-7B-instruct-research-v0.4} & \href{https://huggingface.co/openGPT-X/Teuken-7B-instruct-research-v0.4}{\texttt{openGPT-X/Teuken-7B-instruct-research-v0.4}} \\
European multilingual & \texttt{Pharia-1-LLM-7B-control-hf} & \href{https://huggingface.co/Aleph-Alpha/Pharia-1-LLM-7B-control-hf}{\texttt{Aleph-Alpha/Pharia-1-LLM-7B-control-hf}} \\
European multilingual & \texttt{salamandra-7b-instruct} & \href{https://huggingface.co/BSC-LT/salamandra-7b-instruct}{\texttt{BSC-LT/salamandra-7b-instruct}} \\
Broader multilingual instruction & \texttt{Qwen2.5-7B-Instruct} & \href{https://huggingface.co/Qwen/Qwen2.5-7B-Instruct}{\texttt{Qwen/Qwen2.5-7B-Instruct}} \\
Broader multilingual instruction & \texttt{Llama-3.1-8B-Instruct} & \href{https://huggingface.co/meta-llama/Llama-3.1-8B-Instruct}{\texttt{meta-llama/Llama-3.1-8B-Instruct}} \\
Broader multilingual instruction & \texttt{Mistral-7B-Instruct-v0.3} & \href{https://huggingface.co/mistralai/Mistral-7B-Instruct-v0.3}{\texttt{mistralai/Mistral-7B-Instruct-v0.3}} \\
Broader multilingual instruction & \texttt{gemma-2-9b-it} & \href{https://huggingface.co/google/gemma-2-9b-it}{\texttt{google/gemma-2-9b-it}} \\
Broader multilingual instruction & \texttt{OLMo-2-1124-7B-Instruct} & \href{https://huggingface.co/allenai/OLMo-2-1124-7B-Instruct}{\texttt{allenai/OLMo-2-1124-7B-Instruct}} \\
Broader multilingual instruction & \texttt{granite-3.0-8b-instruct} & \href{https://huggingface.co/ibm-granite/granite-3.0-8b-instruct}{\texttt{ibm-granite/granite-3.0-8b-instruct}} \\
Broader multilingual instruction & \texttt{aya-expanse-8b} & \href{https://huggingface.co/CohereLabs/aya-expanse-8b}{\texttt{CohereLabs/aya-expanse-8b}} \\
\bottomrule
\end{tabular}
}
\caption{Open-weight model targets used for evaluation (\texttt{results\_allmodels} in the repository contains more information). Main result tables use shortened display names for space; this table reports the exact JSON metadata field \texttt{target\_name} and the corresponding model identifier or Hugging Face link where available.}
\label{tab:appendix-model-families}
\end{table*}

The open-weight result repository (\url{https://anonymous.4open.science/r/BavGround-7C68/README.md}) contains 111{,}240 item-level records: 15 model targets, 12 evaluation protocols, and 618 instances per complete model-protocol run. The closed-model directory contains 1{,}236 records for \texttt{gpt-5.4-mini}, covering \texttt{letter} and \texttt{generate\_letter\_and\_text}. The checkpoint directory contains 630{,}360 records: 85 GENBA checkpoint targets, 12 protocols, and 618 instances per checkpoint-protocol run.

Each JSON result file stores run metadata, target identifier, dataset path, filters, generation and diagnostic settings, elapsed time, and per-item outputs. Per-item fields include the item identifier, language, category, prompt, correct option label, parsed prediction, correctness, and protocol-specific diagnostics such as label log-probabilities, generated text, cosine similarities, or hidden-state alignment scores. The semantic matching results use the Sentence Transformers model \texttt{paraphrase-multilingual-mpnet-base-v2}.

The 12 protocols available for the open-weight and checkpoint experiments are \texttt{letter}, \texttt{letter\_shuffled}, \texttt{option\_text}, \texttt{option\_text\_avg}, \texttt{generate\_letter}, \texttt{generate\_option\_text}, \texttt{generate\_letter\_and\_text}, \texttt{semantic\_embed\_generated\_answer}, \texttt{semantic\_embed\_option}, \texttt{semantic\_embed\_contextual\_option}, \texttt{embed\_contextual\_option}, and \texttt{embed\_isolated\_option}. We treat the first eight as output-level evaluation protocols and the final four as semantic or representation-level diagnostics.

We implement the protocol comparison with five scoring families. \textit{Probability-based scoring} includes \texttt{letter}, which scores candidate labels A--D by conditional log-probability, and \texttt{option\_text}/\texttt{option\_text\_avg}, which score the answer texts directly with and without length normalization. \textit{Answer-order perturbation} uses \texttt{letter\_shuffled} to repeat letter scoring after deterministically shuffling the displayed option order. \textit{Generation-based scoring} uses \texttt{generate\_letter}, \texttt{generate\_option\_text}, and \texttt{generate\_letter\_and\_text} to evaluate deterministic generated answers and parse them back to option labels. \textit{External semantic matching} uses the multilingual MPNet sentence-embedding model \texttt{paraphrase-multilingual-mpnet-base-v2}: \texttt{semantic\_embed\_generated\_answer} embeds generated answers and maps them to the closest option, while \texttt{semantic\_embed\_option} and \texttt{semantic\_embed\_contextual\_option} are output-independent question--option baselines. \textit{Hidden-state alignment} compares final-layer hidden states from each evaluated open-weight model using contextual and isolated option representations as representation-level diagnostics rather than token probabilities.

Table~\ref{tab:appendix-protocol-details} summarizes the implementation details that are compressed in the main text. For probability-based open-weight scoring, plain and whitespace-prefixed candidate strings are both considered when tokenization makes the leading-space variant distinct; the better-scoring variant is retained. For margin diagnostics, the correct-answer margin is the score of the correct option minus the highest score assigned to any incorrect option, where incorrect options are the three distractor choices in the multiple-choice item.

\begin{table*}[t]
\centering
\scriptsize
\setlength{\tabcolsep}{3pt}
\resizebox{\textwidth}{!}{
\begin{tabular}{p{3.0cm}p{4.1cm}p{7.1cm}}
\toprule
Protocol family & Strategies & Implementation details \\
\midrule
Letter likelihood & \texttt{letter}; \texttt{letter\_shuffled} & Scores A--D by conditional log-probability. The shuffled variant deterministically permutes displayed option order using the item ID, scores the displayed labels, and maps predictions back to the original option identity. \\
Option-text likelihood & \texttt{option\_text}; \texttt{option\_text\_avg} & Scores the full option text rather than the answer label. \texttt{option\_text} uses summed token log-probability; \texttt{option\_text\_avg} divides by the number of option tokens to reduce option-length effects. \\
Generation & \texttt{generate\_letter}; \texttt{generate\_option\_text}; \texttt{generate\_letter\_and\_text} & Uses deterministic generation with a maximum of 24 new tokens for open-weight runs. Outputs are parsed by initial answer-letter matching, exact option-text matching, contained or prefix option-text matching, and token-overlap matching when needed. \\
External semantic matching & \texttt{semantic\_embed\_option}; \texttt{semantic\_embed\_contextual\_option}; \texttt{semantic\_embed\_generated\_answer} & Uses the multilingual MPNet model named in the main text. The first two strategies are output-independent baselines because they compare the question to option texts without using the evaluated LLM's generated answer; \texttt{semantic\_embed\_generated\_answer} embeds the generated answer and selects the nearest option by cosine similarity. \\
Hidden-state alignment & \texttt{embed\_contextual\_option}; \texttt{embed\_isolated\_option} & Uses the evaluated open-weight model's final-layer hidden states. The final prompt state is the answer anchor and is compared by cosine similarity to mean-pooled option-token states. Contextual alignment embeds options as prompt continuations; isolated alignment embeds options in the template \texttt{Answer: \{option\_text\}}. \\
\bottomrule
\end{tabular}
}
\caption{Evaluation protocol details used by the benchmark runners in \texttt{code/run\_checkpoints\_eval\_v4\_genba\_v2.py}, \texttt{code/run\_checkpoints\_eval\_v4\_genba\_v2\_embed\_v3.py}, and the semantic-embedding job scripts.}
\label{tab:appendix-protocol-details}
\end{table*}

Table~\ref{tab:appendix-bootstrap-ci} reports paired bootstrap intervals for the main comparisons used in the analysis. The bootstrap script is included in the supplementary code as \texttt{code/bootstrap\_biabav\_ci.py}. We resample source question IDs with replacement, so translated English, German, and Bavarian versions of the same source question remain in the same bootstrap cluster. This avoids treating parallel translations as fully independent items.

\begin{table*}[t]
\centering
\small
\setlength{\tabcolsep}{4pt}
\begin{tabular}{p{8.3cm}rr}
\toprule
Comparison & Gap (pp) & 95\% bootstrap CI \\
\midrule
Open-weight mean, \texttt{letter}: GEN $-$ GRD & 16.4 & [8.2, 24.3] \\
Open-weight mean, \texttt{letter}: English $-$ Bavarian & 9.8 & [7.5, 12.2] \\
Open-weight mean, \texttt{letter}: German $-$ Bavarian & 11.7 & [9.5, 14.0] \\
GENBA-10B-it: \texttt{option\_text\_avg} $-$ \texttt{letter} & 17.0 & [10.7, 23.5] \\
\texttt{letter}: EuroLLM-9B $-$ Qwen2.5-7B & 0.3 & [-4.7, 5.3] \\
\texttt{letter}: EuroLLM-9B $-$ Llama-3.1-8B & 0.8 & [-3.2, 4.9] \\
\texttt{letter}: Qwen2.5-7B $-$ Llama-3.1-8B & 0.5 & [-4.2, 5.2] \\
\bottomrule
\end{tabular}
\caption{Nonparametric bootstrap confidence intervals for key accuracy gaps. Intervals use 10{,}000 resamples over source question IDs and are reported in percentage points.}
\label{tab:appendix-bootstrap-ci}
\end{table*}

\section{Additional Open-Weight Model Results}
\label{app:open-model-results}

Table~\ref{tab:appendix-full-letter-model-comparison} provides the full category-expanded version of the main \texttt{letter}-scoring model comparison. The main text reports the compact language and GEN/GRD view for readability.

\begin{table*}[t]
\centering
\scriptsize
\setlength{\tabcolsep}{2.2pt}
\caption{Letter-scoring accuracy (\%) for all evaluated open-weight models on \projname{}, with aggregate language, GEN/GRD, and category breakdowns. Build. = building and sacred heritage; Trad. = living traditions and customs.}
\label{tab:appendix-full-letter-model-comparison}
\resizebox{\textwidth}{!}{
\begin{tabular}{lrrrrrrrrrrrrrr}
\toprule
Model & All & EN & DE & BAV & GEN & GRD & Hist. & Cul. & Lang. & Land. & Pol. & Arts & Build. & Trad. \\
\midrule
EuroLLM-9B & 69.4 & 71.8 & 74.3 & 62.1 & 79.2 & 63.2 & 76.0 & 78.7 & 42.7 & 69.3 & 73.3 & 80.0 & 58.6 & 77.8 \\
Qwen2.5-7B & 69.1 & 72.8 & 67.5 & 67.0 & 74.6 & 65.6 & 70.7 & 80.0 & 48.0 & 66.7 & 73.3 & 77.3 & 65.5 & 71.6 \\
Llama-3.1-8B & 68.6 & 68.4 & 75.2 & 62.1 & 80.4 & 61.1 & 76.0 & 73.3 & 42.7 & 69.3 & 72.0 & 82.7 & 57.5 & 76.5 \\
Granite-3.0-8B & 66.3 & 72.3 & 62.1 & 64.6 & 72.5 & 62.4 & 64.0 & 74.7 & 42.7 & 73.3 & 70.7 & 78.7 & 54.0 & 74.1 \\
Occiglot-7B & 61.8 & 55.8 & 70.9 & 58.7 & 77.5 & 51.9 & 64.0 & 62.7 & 40.0 & 62.7 & 70.7 & 66.7 & 56.3 & 71.6 \\
Aya-8B & 60.8 & 64.6 & 71.4 & 46.6 & 73.3 & 52.9 & 62.7 & 64.0 & 45.3 & 61.3 & 61.3 & 78.7 & 49.4 & 65.4 \\
Mistral-7B & 57.9 & 68.9 & 64.1 & 40.8 & 68.8 & 51.1 & 56.0 & 61.3 & 32.0 & 61.3 & 64.0 & 65.3 & 50.6 & 72.8 \\
OLMo-2-7B & 54.2 & 64.1 & 48.5 & 50.0 & 60.4 & 50.3 & 58.7 & 52.0 & 34.7 & 49.3 & 68.0 & 65.3 & 36.8 & 70.4 \\
Salamandra-7B & 52.1 & 51.9 & 63.1 & 41.3 & 66.7 & 42.9 & 58.7 & 52.0 & 38.7 & 48.0 & 61.3 & 58.7 & 34.5 & 66.7 \\
Gemma-2-9B & 47.7 & 53.9 & 53.9 & 35.4 & 58.8 & 40.7 & 52.0 & 54.7 & 36.0 & 34.7 & 56.0 & 56.0 & 39.1 & 54.3 \\
Leo-7B & 46.4 & 50.5 & 48.5 & 40.3 & 47.5 & 45.8 & 40.0 & 36.0 & 45.3 & 54.7 & 48.0 & 48.0 & 42.5 & 56.8 \\
Teuken-7B & 44.5 & 44.7 & 51.9 & 36.9 & 59.6 & 34.9 & 37.3 & 41.3 & 41.3 & 52.0 & 58.7 & 52.0 & 29.9 & 45.7 \\
Pharia-7B & 43.5 & 49.5 & 48.5 & 32.5 & 62.5 & 31.5 & 38.7 & 45.3 & 26.7 & 50.7 & 56.0 & 48.0 & 32.2 & 51.9 \\
GENBA-10B-it & 41.9 & 35.9 & 50.0 & 39.8 & 55.8 & 33.1 & 46.7 & 32.0 & 14.7 & 52.0 & 52.0 & 50.7 & 34.5 & 53.1 \\
LLaMmlein-7B & 11.2 & 10.2 & 13.6 & 9.7 & 8.8 & 12.7 & 13.3 & 8.0 & 4.0 & 9.3 & 24.0 & 10.7 & 8.0 & 12.3 \\
\midrule
Mean & 53.0 & 55.7 & 57.6 & 45.9 & 63.1 & 46.7 & 54.3 & 54.4 & 35.6 & 54.3 & 60.6 & 61.2 & 43.3 & 61.4 \\
\bottomrule
\end{tabular}
}
\end{table*}

Table~\ref{tab:appendix-allmodel-strategies} reports the full output-level protocol breakdown. The mean score rises from 53.0\% under \texttt{letter} to 59.3\% under \texttt{option\_text\_avg} and 57.1\% under semantic generated-answer matching. The largest \texttt{option\_text\_avg} gains occur for LLaMmlein (+31.7 points), Pharia (+18.9), GENBA (+17.0), Salamandra (+10.7), and Leo (+10.4), showing that answer-format effects are not unique to GENBA.

\begin{table*}[t]
\centering
\scriptsize
\setlength{\tabcolsep}{2.5pt}
\resizebox{\textwidth}{!}{
\begin{tabular}{lrrrrrrrr}
\toprule
Model & Letter & Letter-shuf. & Option-text & Opt.-avg & Gen-letter & Gen-option & Gen-letter+txt & Sem.-gen \\
\midrule
EuroLLM-9B & 69.4 & 69.4 & 60.7 & 68.9 & 69.4 & 76.4 & 52.3 & 54.4 \\
Qwen2.5-7B & 69.1 & 68.0 & 63.3 & 63.9 & 69.1 & 71.8 & 72.5 & 66.7 \\
Llama-3.1-8B & 68.6 & 72.7 & 62.3 & 64.9 & 68.1 & 77.2 & 68.4 & 76.2 \\
Granite-8B & 66.3 & 67.3 & 55.3 & 62.5 & 65.4 & 69.6 & 68.9 & 67.6 \\
Occiglot-7B & 61.8 & 63.9 & 45.6 & 57.6 & 47.9 & 62.6 & 39.5 & 60.4 \\
Aya-Expanse-8B & 60.8 & 62.6 & 54.2 & 58.6 & 64.9 & 70.6 & 61.3 & 70.2 \\
Mistral-7B & 57.9 & 65.0 & 60.4 & 66.5 & 61.2 & 71.2 & 56.8 & 71.2 \\
OLMo-2-7B & 54.2 & 55.8 & 52.1 & 58.6 & 54.0 & 57.9 & 49.5 & 48.7 \\
Salamandra-7B & 52.1 & 57.1 & 53.9 & 62.8 & 57.4 & 72.2 & 66.3 & 69.6 \\
Gemma-2-9B & 47.7 & 53.6 & 42.2 & 55.5 & 73.9 & 75.2 & 74.4 & 50.5 \\
Leo-7B & 46.4 & 40.1 & 42.7 & 56.8 & 41.1 & 46.4 & 32.7 & 31.4 \\
Teuken-7B & 44.5 & 54.2 & 34.3 & 48.4 & 44.5 & 53.7 & 23.1 & 49.0 \\
Pharia-7B & 43.5 & 53.6 & 49.7 & 62.5 & 45.1 & 51.9 & 37.9 & 48.7 \\
GENBA-10B-it & 41.9 & 53.2 & 57.0 & 58.9 & 51.9 & 57.3 & 39.8 & 61.5 \\
LLaMmlein-7B & 11.2 & 24.6 & 20.2 & 42.9 & 0.0 & 0.0 & 0.0 & 30.4 \\
\midrule
Mean & 53.0 & 57.4 & 50.3 & 59.3 & 54.3 & 60.9 & 49.6 & 57.1 \\
\bottomrule
\end{tabular}
}
\caption{Accuracy (\%) across output-level evaluation protocols for all fifteen open-weight models. Opt.-avg is length-normalized option-text scoring; Sem.-gen maps generated answers to options using external semantic similarity.}
\label{tab:appendix-allmodel-strategies}
\end{table*}

Table~\ref{tab:appendix-letter-shuffled-diagnostic} compares canonical \texttt{letter} scoring with \texttt{letter\_shuffled}. The two scores are strongly correlated across models (Pearson $r=0.95$, Spearman $\rho=0.94$; both $p<.001$), and the top four canonical \texttt{letter} models change by only 1.0 point on average. Larger gains are concentrated among lower-scoring or more label-skewed models, so the shuffled diagnostic supports the stability of the main open-model ranking while also showing why protocol-sensitive systems should not be evaluated by canonical \texttt{letter} alone.
    
\begin{table*}[t]
\centering
\small
\setlength{\tabcolsep}{4pt}
\begin{tabular}{lrrr}
\toprule
Model & \texttt{letter} & \texttt{letter\_shuffled} & $\Delta$ \\
\midrule
EuroLLM-9B & 69.4 & 69.4 & +0.0 \\
Qwen2.5-7B & 69.1 & 68.0 & -1.1 \\
Llama-3.1-8B & 68.6 & 72.7 & +4.0 \\
Granite-3.0-8B & 66.3 & 67.3 & +1.0 \\
Occiglot-7B & 61.8 & 63.9 & +2.1 \\
Aya-8B & 60.8 & 62.6 & +1.8 \\
Mistral-7B & 57.9 & 65.0 & +7.1 \\
OLMo-2-7B & 54.2 & 55.8 & +1.6 \\
Salamandra-7B & 52.1 & 57.1 & +5.0 \\
Gemma-2-9B & 47.7 & 53.6 & +5.8 \\
Leo-7B & 46.4 & 40.1 & -6.3 \\
Teuken-7B & 44.5 & 54.2 & +9.7 \\
Pharia-7B & 43.5 & 53.6 & +10.0 \\
GENBA-10B-it & 41.9 & 53.2 & +11.3 \\
LLaMmlein-7B & 11.2 & 24.6 & +13.4 \\
\midrule
Mean & 53.0 & 57.4 & +4.4 \\
\bottomrule
\end{tabular}
\caption{Canonical \texttt{letter} and shuffled-label accuracy (\%) for the 15 open-weight models. \texttt{letter\_shuffled} deterministically permutes displayed answer labels and maps predictions back to original option identities, preserving item content while testing sensitivity to the canonical label-position mapping.}
\label{tab:appendix-letter-shuffled-diagnostic}
\end{table*}

Table~\ref{tab:appendix-letter-label-priors} reports model-specific answer-label distributions under canonical \texttt{letter} scoring. The label-prior expected accuracy is computed from the model's marginal predicted-label distribution and the benchmark's marginal gold-label distribution, without item-level information. The trivial label-only baselines are 25.0\% for uniform random, 9.7\% for always-A, 42.2\% for always-B, 35.9\% for always-C, and 12.1\% for always-D.

\begin{table*}[t]
\centering
\scriptsize
\setlength{\tabcolsep}{2.2pt}
\resizebox{\textwidth}{!}{
\begin{tabular}{lrrrrrrr}
\toprule
Model & Acc. & Pred. A & Pred. B & Pred. C & Pred. D & Label-prior exp. & Residual \\
\midrule
Gold & -- & 60 (9.7) & 261 (42.2) & 222 (35.9) & 75 (12.1) & -- & -- \\
EuroLLM-9B & 69.4 & 159 (25.7) & 189 (30.6) & 166 (26.9) & 104 (16.8) & 27.1 & 42.3 \\
Qwen2.5-7B & 69.1 & 166 (26.9) & 233 (37.7) & 161 (26.1) & 58 (9.4) & 29.0 & 40.1 \\
Llama-3.1-8B & 68.6 & 166 (26.9) & 205 (33.2) & 169 (27.3) & 78 (12.6) & 28.0 & 40.6 \\
Granite-3.0-8B & 66.3 & 144 (23.3) & 196 (31.7) & 198 (32.0) & 80 (12.9) & 28.7 & 37.6 \\
Occiglot-7B & 61.8 & 211 (34.1) & 164 (26.5) & 179 (29.0) & 64 (10.4) & 26.2 & 35.6 \\
Aya-8B & 60.8 & 198 (32.0) & 166 (26.9) & 151 (24.4) & 103 (16.7) & 25.3 & 35.6 \\
Mistral-7B & 57.9 & 274 (44.3) & 150 (24.3) & 132 (21.4) & 62 (10.0) & 23.4 & 34.5 \\
OLMo-2-7B & 54.2 & 220 (35.6) & 231 (37.4) & 119 (19.3) & 48 (7.8) & 27.1 & 27.1 \\
Salamandra-7B & 52.1 & 306 (49.5) & 139 (22.5) & 151 (24.4) & 22 (3.6) & 23.5 & 28.6 \\
Gemma-2-9B & 47.7 & 302 (48.9) & 174 (28.2) & 116 (18.8) & 26 (4.2) & 23.9 & 23.8 \\
Leo-7B & 46.4 & 204 (33.0) & 322 (52.1) & 72 (11.7) & 20 (3.2) & 29.8 & 16.7 \\
Teuken-7B & 44.5 & 259 (41.9) & 114 (18.4) & 90 (14.6) & 155 (25.1) & 20.1 & 24.4 \\
Pharia-7B & 43.5 & 326 (52.8) & 121 (19.6) & 79 (12.8) & 92 (14.9) & 19.8 & 23.7 \\
GENBA-10B-it & 41.9 & 370 (59.9) & 91 (14.7) & 105 (17.0) & 52 (8.4) & 19.2 & 22.8 \\
LLaMmlein-7B & 11.2 & 600 (97.1) & 6 (1.0) & 11 (1.8) & 1 (0.2) & 10.5 & 0.7 \\
\bottomrule
\end{tabular}
}
\caption{Predicted answer-label distributions under canonical \texttt{letter} scoring. Cells for predicted labels report count and percentage. Because the canonical gold-label distribution is imbalanced, raw \texttt{letter} accuracy can be associated with model-specific answer-label priors. Label-prior expected accuracy is computed from the model's marginal predicted-label distribution and the benchmark's marginal gold-label distribution, without item-level information. Residual is actual accuracy minus label-prior expected accuracy. We use this diagnostic together with the shuffled-label comparison in Table~\ref{tab:appendix-letter-shuffled-diagnostic} to distinguish majority-label effects from sensitivity to the canonical label-position mapping.}
\label{tab:appendix-letter-label-priors}
\end{table*}

\begin{table*}[t]
\centering
\small
\setlength{\tabcolsep}{4pt}
\begin{tabular}{lrrp{6.9cm}}
\toprule
Predictor & Pearson $r$ & Spearman $\rho$ & Interpretation \\
\midrule
Predicted B rate & .64 ($p=.010$) & .65 ($p=.009$) & Models that predict B more often tend to score higher, consistent with B being the majority gold label. \\
Predicted C rate & .93 ($p<.001$) & .89 ($p<.001$) & C preference is also strongly associated with accuracy, reflecting that C is the second most frequent gold label. \\
Predicted B+C rate & .89 ($p<.001$) & .77 ($p<.001$) & Preference for the two frequent labels explains substantial between-model variation in raw \texttt{letter} accuracy. \\
Label-prior expected accuracy & .87 ($p<.001$) & .74 ($p=.002$) & Marginal label alignment with the benchmark distribution is strongly associated with leaderboard position. \\
JS distance to gold labels & -.98 ($p<.001$) & -.95 ($p<.001$) & Models whose predicted-label distribution is farther from the gold distribution tend to have lower raw \texttt{letter} accuracy. \\
\bottomrule
\end{tabular}
\caption{Correlations across the 15 open-weight models between actual canonical \texttt{letter} accuracy and answer-label-prior diagnostics. JS denotes Jensen-Shannon distance between the model's predicted-label distribution and the benchmark gold-label distribution. These correlations are descriptive and do not imply that label priors causally determine item-level correctness.}
\label{tab:appendix-letter-label-prior-correlations}
\end{table*}

Table~\ref{tab:appendix-model-metrics} compares all 15 open-weight models across the main evaluation families. Standard letter scoring favours EuroLLM, Qwen, and Llama, while semantic generated-answer matching favours Llama, Mistral, Aya, Salamandra, and Granite. This confirms that model rankings depend on the evaluation view.

\begin{table*}[t]
\centering
\small
\begin{tabular}{lrrrrr}
\toprule
Model & Letter & Opt.-avg & Sem.-gen & H-state ctx. & H-state iso. \\
\midrule
EuroLLM-9B & 69.4 & 68.9 & 54.4 & 27.2 & 28.5 \\
Qwen2.5-7B & 69.1 & 63.9 & 66.7 & 29.9 & 34.5 \\
Llama-3.1-8B & 68.6 & 64.9 & 76.2 & 27.0 & 38.3 \\
Granite-8B & 66.3 & 62.5 & 67.6 & 31.2 & 27.3 \\
Occiglot-7B & 61.8 & 57.6 & 60.4 & 37.1 & 31.2 \\
Aya-Expanse-8B & 60.8 & 58.6 & 70.2 & 23.9 & 37.4 \\
Mistral-7B & 57.9 & 66.5 & 71.2 & 33.0 & 30.9 \\
OLMo-2-7B & 54.2 & 58.6 & 48.7 & 32.8 & 34.8 \\
Salamandra-7B & 52.1 & 62.8 & 69.6 & 26.5 & 33.8 \\
Gemma-2-9B & 47.7 & 55.5 & 50.5 & 13.1 & 30.1 \\
Leo-7B & 46.4 & 56.8 & 31.4 & 30.1 & 34.5 \\
Teuken-7B & 44.5 & 48.4 & 49.0 & 30.6 & 32.2 \\
Pharia-7B & 43.5 & 62.5 & 48.7 & 27.2 & 32.0 \\
GENBA-10B-it & 41.9 & 58.9 & 61.5 & 29.0 & 30.4 \\
LLaMmlein-7B & 11.2 & 42.9 & 30.4 & 14.7 & 50.0 \\
\bottomrule
\end{tabular}
\caption{Open-weight model accuracy (\%) across major evaluation families. Opt.-avg is length-normalized option-text scoring. Sem.-gen maps generated answers to options using external semantic similarity. H-state ctx./iso. are hidden-state alignment diagnostics.}
\label{tab:appendix-model-metrics}
\end{table*}

Tables~\ref{tab:appendix-language-avg} and~\ref{tab:appendix-category-avg} aggregate the open-weight results by language and category. Bavarian is lowest under letter scoring, but the gap narrows under content- and generation-based scoring. Across categories, Language remains the hardest output-level category.

\begin{table*}[t]
\centering
\small
\begin{tabular}{lrrrrr}
\toprule
Category & Letter & Opt.-avg & Sem.-gen & H-state ctx. & H-state iso. \\
\midrule
Arts \& Identity & 61.2 & 61.9 & 67.0 & 30.7 & 40.3 \\
Building \& Sacred & 43.3 & 55.9 & 52.6 & 26.7 & 32.3 \\
Culinary & 54.4 & 62.2 & 60.5 & 27.4 & 27.6 \\
Historical & 54.3 & 68.1 & 57.2 & 22.9 & 33.9 \\
Landscape & 54.3 & 63.5 & 57.2 & 30.7 & 36.7 \\
Language & 35.6 & 39.4 & 45.3 & 30.3 & 27.0 \\
Traditions & 61.4 & 63.3 & 60.5 & 26.5 & 39.8 \\
Politics & 60.6 & 60.2 & 56.9 & 25.6 & 32.1 \\
\bottomrule
\end{tabular}
\caption{Average open-weight model accuracy (\%) by category.}
\label{tab:appendix-category-avg}
\end{table*}

\begin{table*}[t]
\centering
\small
\begin{tabular}{lrrr}
\toprule
Metric & EN & DE & BAR \\
\midrule
Letter & 55.7 & 57.6 & 45.9 \\
Option-text-avg & 61.9 & 59.8 & 56.1 \\
Semantic generated & 56.0 & 59.7 & 55.7 \\
Hidden-state contextual & 29.5 & 26.5 & 26.6 \\
Hidden-state isolated & 33.6 & 33.5 & 34.1 \\
\bottomrule
\end{tabular}
\caption{Average open-weight model accuracy (\%) by language.}
\label{tab:appendix-language-avg}
\end{table*}

\section{Closed-Model Reference Results}
\label{app:closed-model-results}

The closed-model run in \texttt{results\_closedmodels} evaluates the OpenAI API model identifier \texttt{gpt-5.4-mini} under \texttt{letter} and \texttt{generate\_letter\_and\_text}. The stored result files are timestamped 2026-05-12 12:11:31. For \texttt{letter}, we score A--D using first-token API log-probabilities with \texttt{top\_logprobs=5}; for \texttt{generate\_letter\_and\_text}, we parse generated answers from a prompt asking for the letter followed by optional option text. The runner configuration stores \texttt{temperature=0}, \texttt{temperature\_sent\_to\_api=None}, \texttt{letter\_max\_completion\_tokens=64}, and \texttt{generate\_max\_completion\_tokens=128}. The artifacts do not include a more specific dated snapshot suffix beyond the API model ID, so we report the model ID and run timestamp rather than treating the closed model as a fixed open-weight checkpoint.

\begin{table*}[t]
\centering
\small
\setlength{\tabcolsep}{2.5pt}
\resizebox{\textwidth}{!}{
\begin{tabular}{lrrrrrrrrrrrr}
\toprule
Strategy & All & EN & DE & BAR & Arts & Build. & Cul. & Hist. & Land. & Lang. & Trad. & Pol. \\
\midrule
Letter & 89.6 & 89.3 & 91.3 & 88.3 & 97.3 & 79.3 & 94.7 & 88.0 & 90.7 & 81.3 & 97.5 & 89.3 \\
Generate-letter+txt & 88.0 & 89.8 & 88.8 & 85.4 & 96.0 & 83.9 & 93.3 & 84.0 & 90.7 & 77.3 & 93.8 & 85.3 \\
\bottomrule
\end{tabular}
}
\caption{\texttt{gpt-5.4-mini} accuracy (\%) by language and category under the two closed-model protocols available in \texttt{results\_closedmodels}. Build. = building and sacred heritage; Trad. = living traditions and customs; Pol. = politics.}
\label{tab:appendix-closed-model}
\end{table*}

\section{GENBA Protocol Diagnostics}
\label{app:genba-protocol-diagnostics}

\begin{table*}[t]
\centering
\small
\setlength{\tabcolsep}{3pt}
\caption{GENBA-10B-it accuracy (\%) under different evaluation protocols. Sem.-gen denotes semantic matching of generated answers.}
\label{tab:appendix-genba-protocol-summary}
\begin{tabular}{lrrrrrr}
\toprule
Strategy & All & EN & DE & BAV & GEN & GRD \\
\midrule
Letter              & 41.9 & 35.9 & 50.0 & 39.8 & 55.8 & 33.1 \\
Letter-shuf.        & 53.2 & 48.5 & 60.2 & 51.0 & 63.3 & 46.8 \\
Option-text         & 57.0 & 61.2 & 56.8 & 52.9 & 70.0 & 48.7 \\
Option-text-avg     & 58.9 & \textbf{63.6} & 59.2 & 53.9 & 67.1 & \textbf{53.7} \\
Generate-letter     & 51.9 & 46.6 & 60.2 & 49.0 & 64.2 & 44.2 \\
Generate-option     & 57.3 & 57.8 & 63.1 & 51.0 & 72.5 & 47.6 \\
Generate-letter+txt & 39.8 & 38.8 & 45.1 & 35.4 & 55.0 & 30.2 \\
Sem.-gen            & \textbf{61.5} & 62.6 & \textbf{65.0} & \textbf{56.8} & \textbf{73.8} & \textbf{53.7} \\
\bottomrule
\end{tabular}
\end{table*}

Table~\ref{tab:appendix-genba-strategies} gives the full GENBA strategy breakdown. GENBA improves from 41.9\% under letter scoring to 58.9\% under length-normalized option-text scoring and 61.5\% under semantic matching of generated answers. This supports the main claim that standard answer-letter evaluation understates the amount of answer content recovered by alternative protocols.

\begin{table*}[t]
\centering
\small
\begin{tabular}{lrrrr}
\toprule
Strategy & All & EN & DE & BAR \\
\midrule
Letter & 41.9 & 35.9 & 50.0 & 39.8 \\
Letter-shuffled & 53.2 & 48.5 & 60.2 & 51.0 \\
Option-text & 57.0 & 61.2 & 56.8 & 52.9 \\
Option-text-avg & 58.9 & 63.6 & 59.2 & 53.9 \\
Generate-letter & 51.9 & 46.6 & 60.2 & 49.0 \\
Generate-option & 57.3 & 57.8 & 63.1 & 51.0 \\
Generate-letter+text & 39.8 & 38.8 & 45.1 & 35.4 \\
Semantic generated & 61.5 & 62.6 & 65.0 & 56.8 \\
Hidden-state contextual & 29.0 & 32.5 & 25.7 & 28.6 \\
Hidden-state isolated & 30.4 & 28.2 & 30.1 & 33.0 \\
\bottomrule
\end{tabular}
\caption{GENBA-10B-it accuracy (\%) across evaluation strategies and languages.}
\label{tab:appendix-genba-strategies}
\end{table*}

The protocol effect is category-dependent (Table~\ref{tab:appendix-genba-category-strategies}). Historical questions show the largest recovery under \texttt{option\_text\_avg}, rising from 46.7\% to 80.0\%. Language rises from 14.7\% under \texttt{letter} to 56.0\% under \texttt{option\_text\_avg}, but remains weaker than most other categories under several generation protocols.

\begin{table*}[t]
\centering
\small
\setlength{\tabcolsep}{2.5pt}
\resizebox{\textwidth}{!}{
\begin{tabular}{lrrrrrrrr}
\toprule
Category & Letter & Letter-shuf. & Option-text & Opt.-avg & Gen-letter & Gen-option & Gen-letter+txt & Sem.-gen \\
\midrule
Historical & 46.7 & 53.3 & 69.3 & 80.0 & 57.3 & 70.7 & 44.0 & 72.0 \\
Culinary & 32.0 & 49.3 & 46.7 & 54.7 & 42.7 & 54.7 & 28.0 & 66.7 \\
Language & 14.7 & 36.0 & 65.3 & 56.0 & 22.7 & 36.0 & 10.7 & 48.0 \\
Landscape & 52.0 & 60.0 & 61.3 & 60.0 & 65.3 & 65.3 & 45.3 & 66.7 \\
Politics & 52.0 & 57.3 & 58.7 & 56.0 & 54.7 & 56.0 & 54.7 & 50.7 \\
Arts \& Identity & 50.7 & 53.3 & 60.0 & 56.0 & 66.7 & 68.0 & 52.0 & 69.3 \\
Building \& Sacred & 34.5 & 57.5 & 43.7 & 55.2 & 47.1 & 46.0 & 36.8 & 54.0 \\
Traditions & 53.1 & 58.0 & 53.1 & 54.3 & 59.3 & 63.0 & 46.9 & 65.4 \\
\bottomrule
\end{tabular}
}
\caption{GENBA-10B-it category-level accuracy (\%) across output-level evaluation protocols.}
\label{tab:appendix-genba-category-strategies}
\end{table*}

Table~\ref{tab:appendix-answer-distribution} shows the answer-label distributions behind this protocol sensitivity. GENBA heavily overpredicts option A under standard \texttt{letter} scoring and \texttt{generate\_letter\_and\_text}, while shuffled labels, option-text scoring, and semantic matching produce more balanced distributions.

\begin{table*}[h]
\centering
\small
\begin{tabular}{lrrrr}
\toprule
Strategy & A & B & C & D \\
\midrule
Gold & 60 & 261 & 222 & 75 \\
Letter & 370 & 91 & 105 & 52 \\
Letter-shuffled & 130 & 201 & 174 & 113 \\
Option-text-avg & 151 & 228 & 146 & 93 \\
Generate-letter+text & 417 & 84 & 95 & 22 \\
Semantic generated & 185 & 172 & 163 & 98 \\
\bottomrule
\end{tabular}
\caption{GENBA-10B-it predicted answer-label distributions across selected strategies.}
\label{tab:appendix-answer-distribution}
\end{table*}

\section{Additional Checkpoint Results}
\label{app:checkpoint-results}

The checkpoint directory contains 85 GENBA checkpoints from checkpoint 0 through checkpoint 41{,}707. Table~\ref{tab:appendix-checkpoint-metrics} summarizes trajectories by protocol. \texttt{option\_text\_avg} improves from 25.4\% to 50.8\%, peaking at 51.3\% near checkpoint 41.5k. In contrast, letter accuracy peaks much lower, at 19.3\% around checkpoint 18.5k, and ends at 12.1\%.

\begin{table*}[t]
\centering
\small
\begin{tabular}{lrrrr}
\toprule
Metric & Ckpt 0 & Best Ckpt & Best & Final \\
\midrule
Letter & 11.2 & 18.5k & 19.3 & 12.1 \\
Option-text & 16.8 & 16k & 40.3 & 38.3 \\
Option-text-avg & 25.4 & 41.5k & 51.3 & 50.8 \\
Generate-letter & 11.8 & 14.5k & 37.7 & 23.0 \\
Generate-option & 11.2 & 14.5k & 44.0 & 29.4 \\
Generate-letter+text & 11.7 & 14.5k & 29.8 & 17.0 \\
Semantic generated & 27.2 & 14.5k & 50.3 & 41.9 \\
Hidden-state contextual & 25.9 & 27.5k & 29.8 & 29.3 \\
Hidden-state isolated & 35.6 & 0 & 35.6 & 29.9 \\
\bottomrule
\end{tabular}
\caption{GENBA checkpoint trajectories by evaluation metric. Final refers to checkpoint 41{,}707.}
\label{tab:appendix-checkpoint-metrics}
\end{table*}

Table~\ref{tab:appendix-checkpoint-error-examples} illustrates the kinds of errors behind the metric divergence at checkpoint 14.5k. The examples are drawn directly from the stored checkpoint outputs. They show three recurring patterns: a strong prior for option A under strict letter parsing, generated text that contains the correct answer content but does not begin with the correct parseable label, and dialect expressions where option-text likelihood identifies the correct meaning while both letter generation and contextual hidden-state alignment select a distractor. For log-probability scores, less negative values indicate stronger model preference.

\begin{table*}[t]
\centering
\scriptsize
\setlength{\tabcolsep}{3pt}
\resizebox{\textwidth}{!}{
\begin{tabular}{p{2.1cm}p{2.7cm}p{3.4cm}p{4.3cm}p{4.2cm}}
\toprule
Item & Question focus & Gold and wrong option & Letter / generation evidence & Recovered or conflicting signal \\
\midrule
\texttt{culinary\_en\_012} & Cultural meaning of \textit{saure Lüngerl} & Gold B: humble dish served at weddings and funeral meals in scarcity. Wrong A: festive dish for wealthy families. & Letter prefers A over B (A $-2.41$, B $-3.97$). Generated letter+text produces an unfinished explanation without a parseable correct label. & Option-text-avg prefers B (B $-0.27$, C $-0.63$, A $-0.80$, D $-1.18$). Semantic similarities also prefer B (B .69, A .58). \\
\texttt{historical\_de\_011} & Ludwig III leaving Munich on 7 Nov. 1918 & Gold C: chauffeur joined the revolution and official cars were unusable. Wrong A: loyal troops had surrounded the city. & Letter prefers A over C (A $-2.39$, C $-2.89$). Generated letter+text begins by listing distractor options and is parsed as A. & Option-text-avg narrowly prefers C over A (C $-0.52$, A $-0.54$). Semantic generation strongly prefers C (C .96 vs. B .28, A .21). Contextual hidden-state alignment also selects C. \\
\texttt{language\_en\_001} & Meaning of ``Geh, hör auf!'' & Gold B: Stop it / You're kidding me! Wrong A: Go, listen up! & Letter strongly prefers A over B (A $-2.48$, B $-4.34$). Generated letter+text outputs ``A: Go, listen up!'' & Option-text-avg recovers B (B $-1.01$, A $-1.15$, C $-1.33$, D $-4.36$). Semantic generation and contextual hidden-state alignment still select A, reflecting an overly literal interpretation. \\
\bottomrule
\end{tabular}
}
\caption{Representative checkpoint-14.5k output patterns illustrating why checkpoint metrics diverge. Letter scores are label log-probabilities; option-text-avg scores are average option-text log-probabilities; semantic values are cosine similarities. For log-probabilities, less negative is better.}
\label{tab:appendix-checkpoint-error-examples}
\end{table*}

Tables~\ref{tab:appendix-checkpoint-domain} and~\ref{tab:appendix-checkpoint-language} summarize \texttt{option\_text\_avg} trajectories by category and language. Language remains the weakest final category at 30.7\%, while Bavarian remains below English and German at the final checkpoint.

\begin{table*}[h]
\centering
\small
\begin{tabular}{lrrrr}
\toprule
Domain & Ckpt 0 & Best Ckpt & Best & Final \\
\midrule
Arts \& Identity & 21.3 & 39k & 60.0 & 58.7 \\
Building \& Sacred & 39.1 & 23k & 59.8 & 51.7 \\
Culinary & 28.0 & 28k & 64.0 & 57.3 \\
Historical & 36.0 & 41.5k & 58.7 & 53.3 \\
Landscape & 28.0 & 16k & 58.7 & 53.3 \\
Language & 18.7 & 33k & 30.7 & 30.7 \\
Traditions & 22.2 & 6k & 56.8 & 53.1 \\
Politics & 8.0 & 4.5k & 53.3 & 48.0 \\
\bottomrule
\end{tabular}
\caption{GENBA domain trajectories under \texttt{option\_text\_avg}. Final refers to checkpoint 41{,}707.}
\label{tab:appendix-checkpoint-domain}
\end{table*}

\begin{table*}[h]
\centering
\small
\begin{tabular}{lrrrr}
\toprule
Subset & Ckpt 0 & Best Ckpt & Best & Final \\
\midrule
All & 25.4 & 41.5k & 51.3 & 50.8 \\
EN & 29.1 & 23.5k & 57.3 & 54.4 \\
DE & 25.2 & 23k & 57.3 & 56.3 \\
BAR & 21.8 & 38.5k & 44.7 & 41.7 \\
\bottomrule
\end{tabular}
\caption{GENBA checkpoint accuracy (\%) by language under \texttt{option\_text\_avg}. Final refers to checkpoint 41{,}707.}
\label{tab:appendix-checkpoint-language}
\end{table*}

\begin{figure*}[t]
    \centering
    \includegraphics[width=\textwidth]{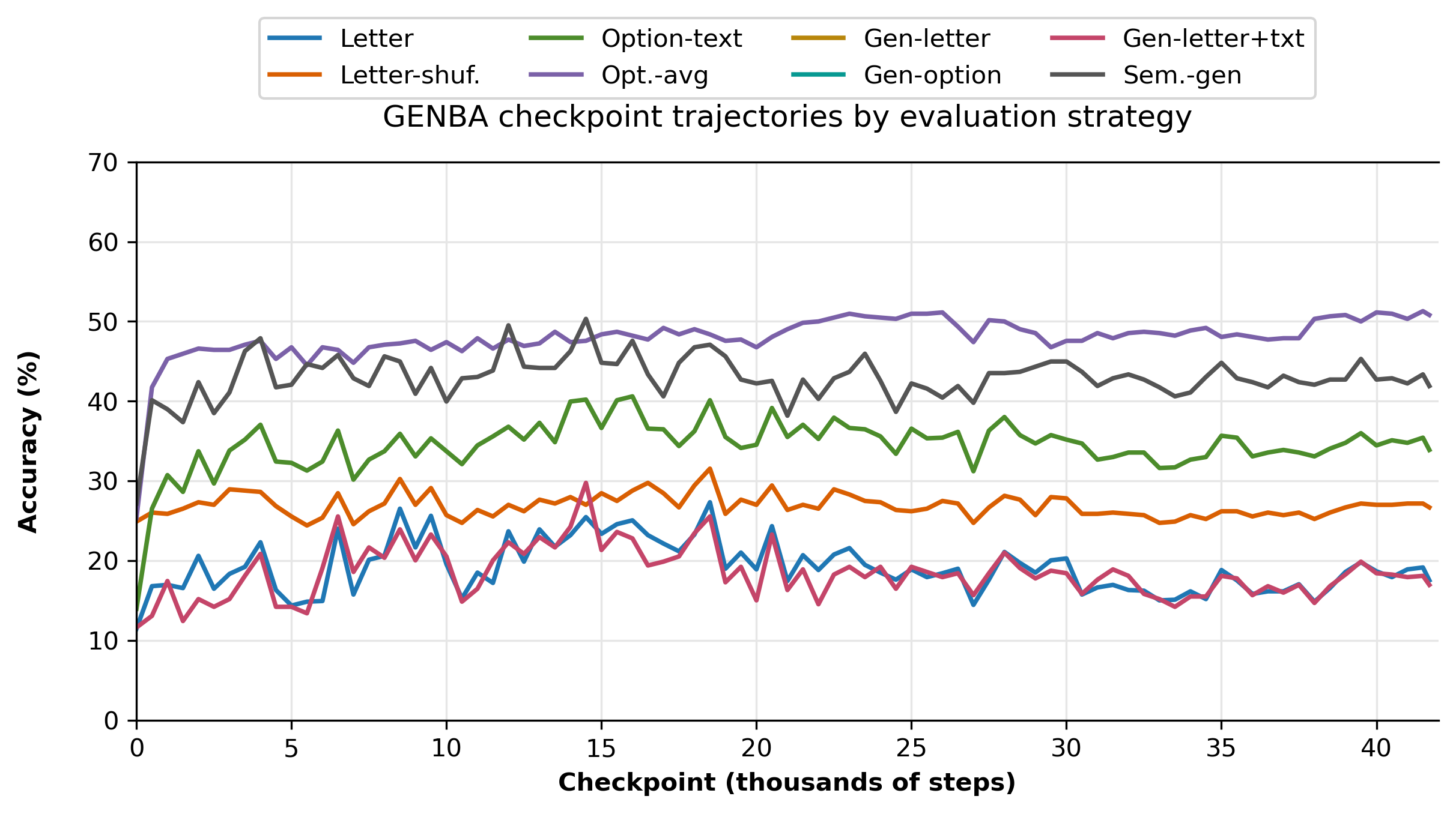}
    \caption{GENBA-10B checkpoint trajectories by evaluation strategy, aggregated across all languages and categories. This figure complements the two-line main-text plot by showing the remaining output-level protocols.}
    \label{fig:checkpoint-strategy-trajectories}
\end{figure*}

\begin{figure*}[t]
    \centering
    \includegraphics[width=\textwidth]{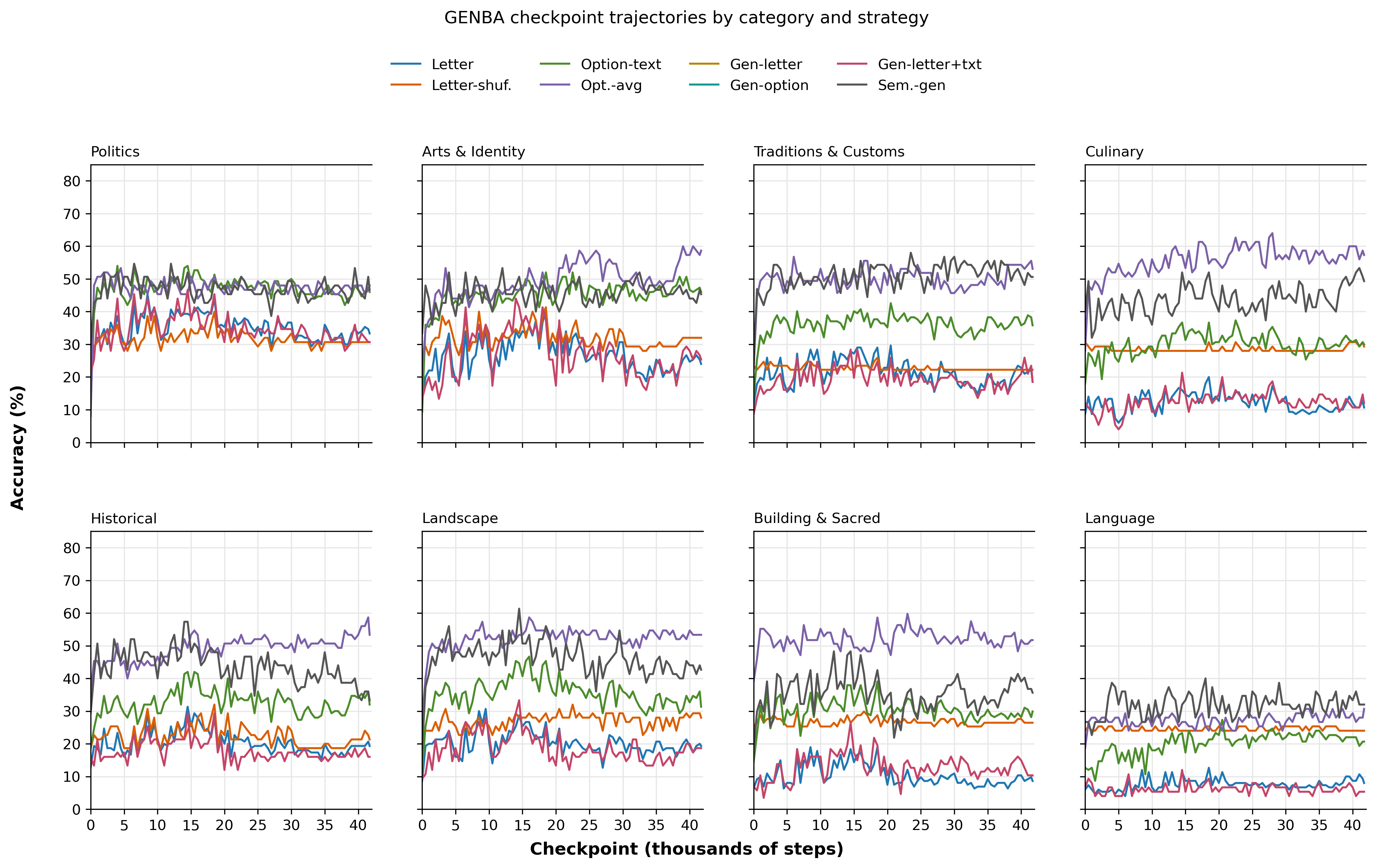}
    \caption{GENBA-10B checkpoint trajectories by category and evaluation strategy, using all available checkpoints. Each panel corresponds to one \projname{} category and each line corresponds to one output-level evaluation protocol.}
    \label{fig:checkpoint-category-strategy-trajectories}
\end{figure*}

\begin{figure*}[t]
    \centering
    \includegraphics[width=\textwidth]{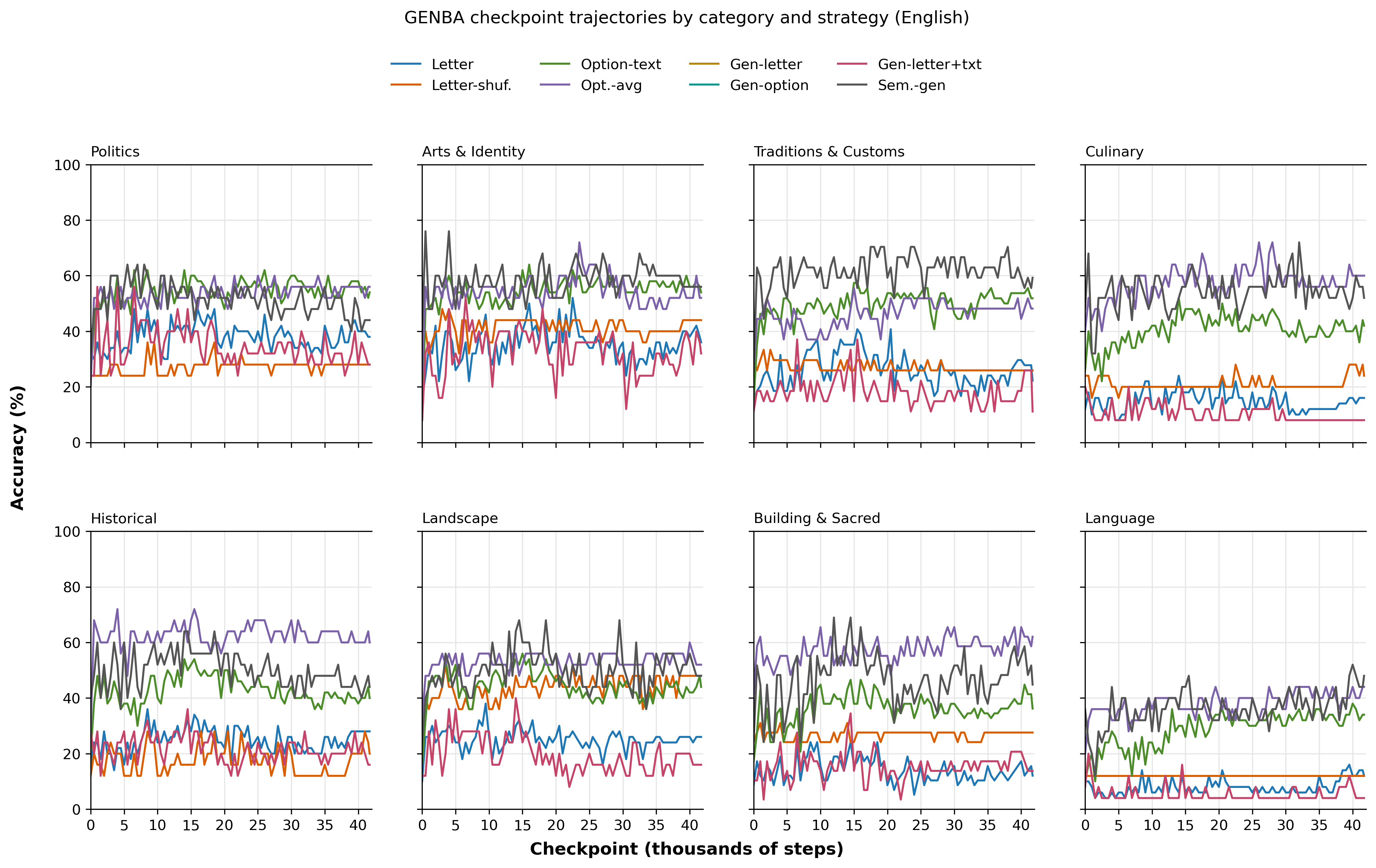}
    \caption{GENBA-10B checkpoint trajectories for English items by category and evaluation strategy.}
    \label{fig:checkpoint-en-category-strategy-trajectories}
\end{figure*}

\begin{figure*}[t]
    \centering
    \includegraphics[width=\textwidth]{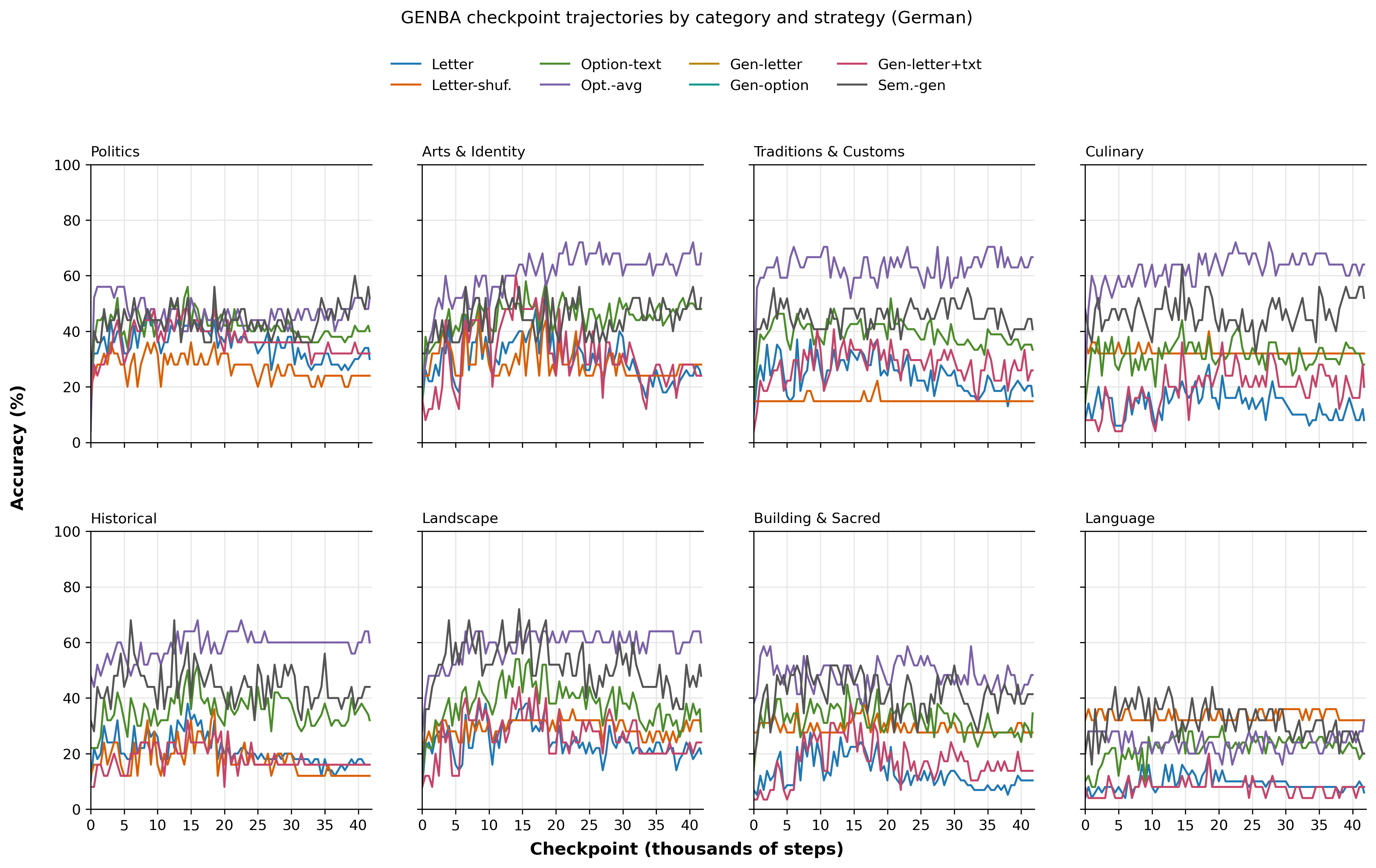}
    \caption{GENBA-10B checkpoint trajectories for German items by category and evaluation strategy.}
    \label{fig:checkpoint-de-category-strategy-trajectories}
\end{figure*}

\begin{figure*}[t]
    \centering
    \includegraphics[width=\textwidth]{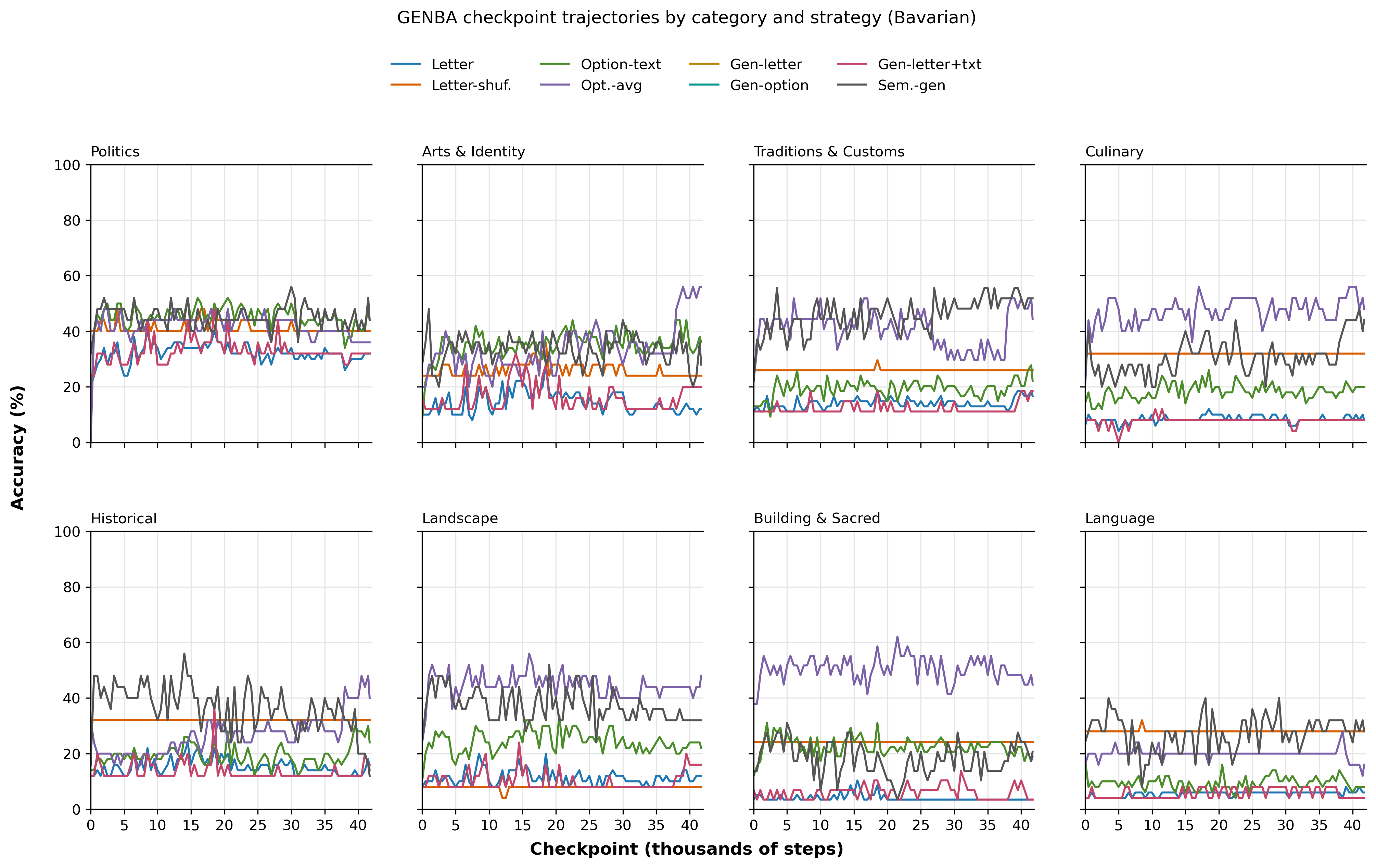}
    \caption{GENBA-10B checkpoint trajectories for Bavarian items by category and evaluation strategy.}
    \label{fig:checkpoint-bar-category-strategy-trajectories}
\end{figure*}

Figures~\ref{fig:checkpoint-en-category-strategy-trajectories}--\ref{fig:checkpoint-bar-category-strategy-trajectories} separate the category-strategy trajectories by language. These figures show that aggregate checkpoint improvements can hide language-specific instability, especially for Bavarian items.

\section{Interactive Analysis Dashboard}
\label{app:dashboard}

\begin{figure*}[t]
    \centering
    \includegraphics[width=\textwidth]{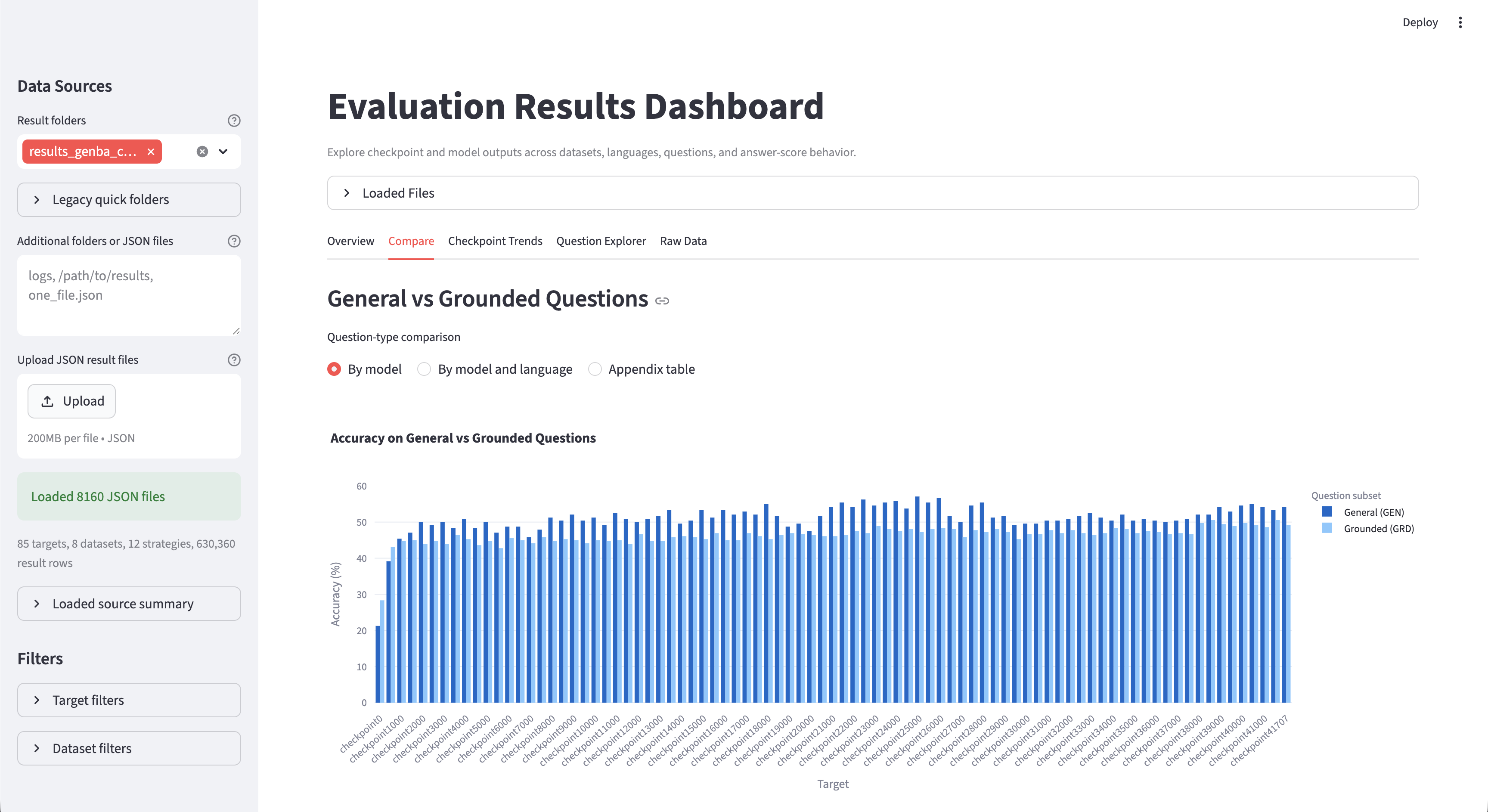}
    \caption{Interactive dashboard for inspecting \projname{} evaluation outputs across models, checkpoints, languages, domains, and scoring strategies.}
    \label{fig:dashboard}
\end{figure*}

We include the Streamlit dashboard code in the supplementary material. The dashboard entry point is \texttt{dashboard/app.py}; it loads the same per-item JSON files summarized above and supports filtering by target model or checkpoint, language, category, question subset, and evaluation strategy. It visualizes aggregate accuracy, category-level accuracy, checkpoint trajectories, and answer-choice distributions, making it useful for error analysis and reproducibility rather than as a separate metric. Figure~\ref{fig:dashboard} shows the dashboard interface.

\end{document}